\documentclass{article}

    \usepackage[final]{neurips_2024}

\usepackage[utf8]{inputenc} 
\usepackage[T1]{fontenc}    
\usepackage{hyperref}       
\usepackage{url}            
\usepackage{booktabs}       
\usepackage{amsfonts}       
\usepackage{nicefrac}       
\usepackage{microtype}      
\usepackage{xcolor}         
\usepackage{natbib}
\setcitestyle{numbers,open={[},close={]}} 
\usepackage{graphicx}
\usepackage{multirow}
\usepackage{subfigure}
\usepackage{enumitem}
\usepackage{amsmath}
\usepackage{float}
\usepackage{colortbl}
\usepackage{algorithm}
\usepackage{algpseudocode}
\usepackage[compact]{titlesec}

\title{LLMs as Zero-shot Graph Learners: Alignment of GNN Representations with LLM Token Embeddings}

\author{
  Duo Wang \quad Yuan Zuo\thanks{Corresponding author.} \quad Fengzhi Li \quad Junjie Wu \\
  MIIT Key Laboratory of Data Intelligence and Management, Beihang University \\
  \{wangduo58, zuoyuan, lifengzhi, wujj\}@buaa.edu.cn \\
}

\begin{document}

\maketitle





\begin{abstract}
Zero-shot graph machine learning, especially with graph neural networks (GNNs), has garnered significant interest due to the challenge of scarce labeled data. While methods like self-supervised learning and graph prompt learning have been extensively explored, they often rely on fine-tuning with task-specific labels, limiting their effectiveness in zero-shot scenarios. Inspired by the zero-shot capabilities of instruction-fine-tuned large language models (LLMs), we introduce a novel framework named Token Embedding-Aligned Graph Language Model (TEA-GLM) that leverages LLMs as cross-dataset and cross-task zero-shot learners for graph machine learning. Concretely, we pretrain a GNN, aligning its representations with token embeddings of an LLM. We then train a linear projector that transforms the GNN's representations into a fixed number of graph token embeddings without tuning the LLM. A unified instruction is designed for various graph tasks at different levels, such as node classification (node-level) and link prediction (edge-level). These design choices collectively enhance our method's effectiveness in zero-shot learning, setting it apart from existing methods. Experiments show that our graph token embeddings help the LLM predictor achieve state-of-the-art performance on unseen datasets and tasks compared to other methods using LLMs as predictors. Our code is available at~\url{https://github.com/W-rudder/TEA-GLM}.
\end{abstract}

\section{Introduction}
\label{sect:intro}
Graph Neural Networks (GNNs) have emerged as a pivotal framework in graph machine learning, harnessing the ability to capture intricate message-passing patterns for robust graph representation. These advancements have yielded various GNN architectures, including the Graph Convolution Network (GCN)~\cite{GCN}, Graph Attention Network (GAT)~\cite{GAT}, and GraphSAGE~\cite{SAGE}. Despite their efficacy, GNNs often exhibit limited generalization capabilities, struggling to maintain consistent performance when transitioning across different datasets or downstream tasks~\cite{ParetoGNN}. This limitation underscores the necessity for more adaptable and universally applicable models in the graph learning domain.

To mitigate the dependency on labeled data and enhance the resilience of graph models, self-supervised learning has been widely adopted in GNN training. Techniques such as Deep Graph Infomax (DGI)~\cite{DGI} and GraphCL~\cite{GraphCL} have demonstrated effectiveness by leveraging mutual information maximization and contrastive learning, respectively. However, these methods typically require fine-tuning task-specific heads for downstream applications, which can be resource-intensive and limit their practicality in diverse scenarios. Moreover, graph prompt learning enhances GNN generalization by using unified task templates and meta-learning to adapt to various downstream applications~\cite{graphprompt, allinone}, but it often requires extensive fine-tuning and is constrained by the specificity of task types.

In recent years, the remarkable generalization capabilities of Large Language Models (LLMs) have spurred interest in their potential applications within graph machine learning. Some methods attempt to encode graph structures into text for LLM input~\cite{graph2text, gpt4graph, nlgraph, llm2graph}, but these approaches often lead to suboptimal outcomes~\cite{canllm}. Alternatively, using LLMs as enhancers to generate data or node text representations~\cite{natureisneed, llmgnn1, llmgnn2, labelfree, OFA} has shown promise but remains constrained by the inherent reliance on GNNs for prediction. Recent efforts~\cite{graphgpt, llaga} to use LLMs as predictors have demonstrated potential. However, their performance often remains unstable due to the challenge of producing transferable graph representations that work effectively for LLMs across diverse tasks and datasets.

In light of these challenges, we propose a novel framework named Token Embedding-Aligned Graph Language Model (TEA-GLM). Inspired by the zero-shot capabilities of instruction-fine-tuned LLMs~\cite{finetunedlmforzs}, TEA-GLM leverages LLMs as cross-dataset and cross-task zero-shot predictors for graph machine learning. The core idea is to pretrain a GNN and align its representations with the token embeddings of an LLM. This alignment enables the GNN to effectively utilize the LLM's pretrained knowledge, allowing it to generalize across different datasets and tasks without task-specific fine-tuning. Additionally, we train a linear projector to convert graph representations into a fixed number of token embeddings, which are then incorporated into a unified instruction designed for various graph tasks at different levels. Experiments show TEA-GLM achieves superior performance in zero-shot scenarios and when encountering unseen tasks, offering a more generalized and efficient solution for graph zero-shot learning. Our contributions are summarized as follows:
\begin{itemize}[leftmargin=8pt]
    \item We introduce TEA-GLM, a novel framework that aligns GNN representations with LLM token embeddings, enabling cross-dataset and cross-task zero-shot learning for graph machine learning.
    \item We propose a linear projector that maps graph representations into a fixed number of graph token embeddings. These embeddings are incorporated into a unified instruction designed for various graph tasks at different levels, enhancing the model's generalization capabilities.
    \item Our extensive experiments demonstrate that TEA-GLM significantly outperforms state-of-the-art methods on unseen datasets and tasks.
\end{itemize}

\section{Methodology}
\label{sect:method}
\begin{figure}[t!]
    \centering
    \includegraphics[scale=0.38]{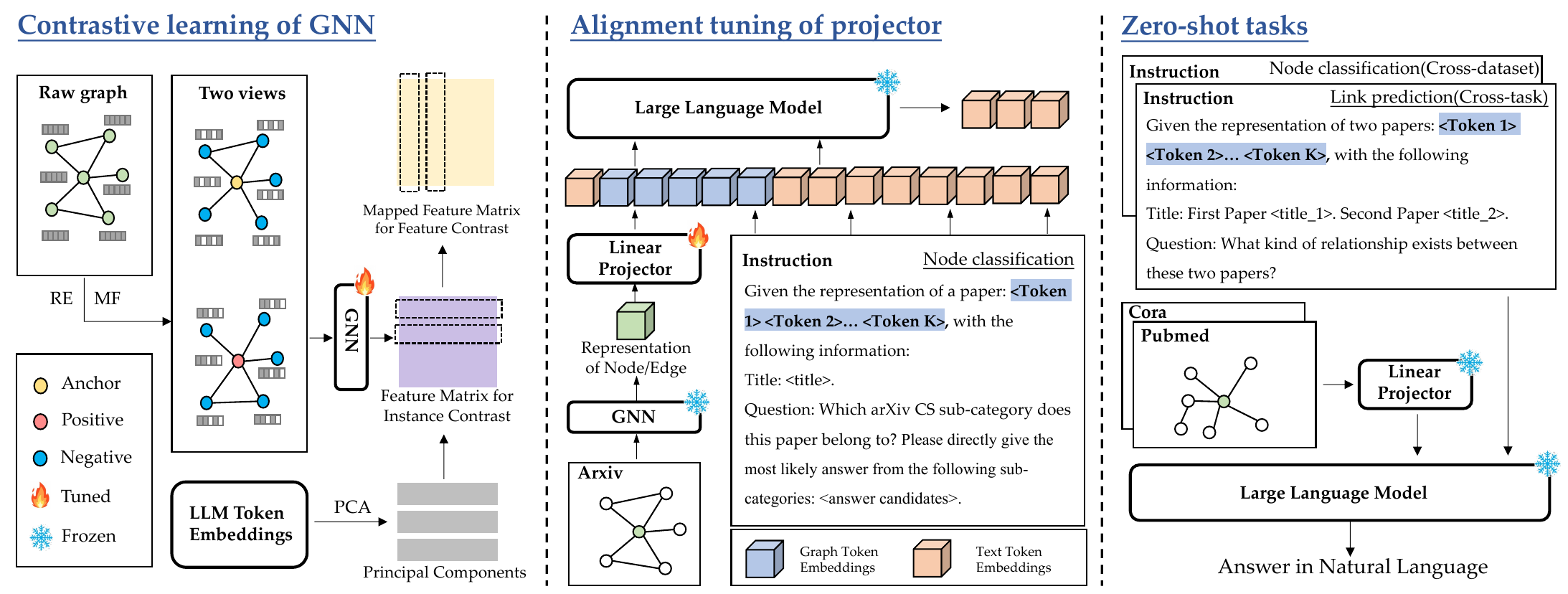}
        \caption{Framework of TEA-GLM}
    \label{fig:framework}
\end{figure}

In this section, we introduce TEA-GLM, a novel framework designed for cross-dataset and cross-task zero-shot graph machine learning. TEA-GLM consists of two main components: a Graph Neural Network (GNN) to derive node representations from the graph, and a Large Language Model (LLM) to perform zero-shot tasks such as node classification and link prediction. Our methodology involves two key stages: enhanced self-supervised learning of the GNN, where feature-wise contrastive learning with LLM's token embeddings is proposed, and training a linear projector to map graph representations into a fixed number of graph token embeddings by designing an instruction that is suitable for various graph tasks at different levels. The framework of our proposed method is illustrated in Fig.~\ref{fig:framework}.

\subsection{Notations}

Formally, a graph is denoted as $\mathcal{G} = \left ( \mathcal{V}, \mathcal{E}, \mathbf{A}, \mathbf{X} \right ) $, where $\mathcal{V} = \left \{ v_1, v_2, \dots , v_{\left | \mathcal{V} \right |}  \right \} $ with $\left | \mathcal{V}  \right | = N$ indicating the total number of nodes and $\mathcal{E} = \left \{ e_1, e_2, \dots , e_{\left | \mathcal{E} \right |} \right \}$ representing the sets of nodes and edges, respectively. The adjacency matrix is denoted as $\mathbf{A} \in \mathbb{R} ^{N \times N } $, with $\mathbf{A}_{ij} = 1$ iff $(v_i, v_j) \in \mathcal{E}$. The feature matrix $\mathbf{X} \in \mathbb{R} ^{N \times F_N}$ contains the attribute or feature information associated with each node, where $\boldsymbol{x_i} \in \mathbb{R}^{F_N}$ is the feature of ${v_i}$, and $F_N$ represents the dimensionality of features.

\subsection{Token embeddings-aligned graph self-supervised learning}


Given the increasing model sizes and data volumes in recent years, self-supervised learning has become a prominent research focus due to the scarcity of labeled data. In this context, we propose a contrastive learning method to obtain more transferable node representations suitable for use with large language models (LLMs). Our approach leverages instance-wise contrastive learning and introduces a feature-wise contrastive learning method that maps node representations to the textual embedding space of the LLM.

\subsubsection{Instance-wise contrastive learning with structural information}

To alleviate the need for labeled data and enhance model generalization capability, we employ self-supervised learning for pre-training. To better extract structural information from the graph, we follow the work of \cite{zhu2020deep} to generate two views of $\mathcal{G}$, denoted as ${\mathcal{G}_{1}}$ and ${\mathcal{G}_{2}}$, for contrastive learning. Specifically, we adopt the Removing Edges (RE) and Masking Node Features (MF) methods to generate different views. The RE strategy samples a random masking matrix $\widetilde{\mathbf{R}} \in \left \{ 0, 1 \right \} ^{N \times N }$ to mask the raw adjacency matrix, computed as:
\begin{equation}
    \widetilde{\mathbf{A}} = \mathbf{A} \circ \widetilde{\mathbf{R}},
\end{equation}
where $\circ$ denotes the Hadamard product. The MF strategy samples a random mask vector $\widetilde{\boldsymbol{m}} \in \left \{ 0, 1 \right \} ^F$. The generated node features $\widetilde{\mathbf{X}}$ are computed by:
\begin{equation}
    \widetilde{\mathbf{X}} = \left[ \boldsymbol{x_1} \circ \widetilde{\boldsymbol{m}}; \boldsymbol{x_2} \circ \widetilde{\boldsymbol{m}}; \ldots; \boldsymbol{x_N} \circ \widetilde{\boldsymbol{m}} \right].
\end{equation}
Thus, we obtain two views of $\mathcal{G}$, denoted as ${\mathcal{G}_{1}} = \left( \widetilde{\mathbf{X}}_1, \widetilde{\mathbf{A}}_1\right)$ and ${\mathcal{G}_{2}} = \left( \widetilde{\mathbf{X}}_2, \widetilde{\mathbf{A}}_2\right)$. Then, we use a graph encoder to derive node representations:
\begin{equation}
    \mathbf{U_\ast} = f_\mathrm{GNN}\left(\widetilde{\mathbf{X}}_\ast, \widetilde{\mathbf{A}}_\ast \right) \in \mathbb{R}^{N \times F_U},
\end{equation}
Where $F_U$ is the dimension size of node representations. Here, $\ast \in \left\{ 1, 2\right\}$ represents different views of the graph.

We employ a contrastive objective to distinguish the embeddings of the same node in these two different views from other node embeddings. For node $v_i$, its node embedding generated in one view, $\boldsymbol{u_i}$, is treated as the anchor, while the embedding generated in the other view, $\boldsymbol{u_i}^\prime$, forms the positive sample. Embeddings of other nodes in the same view are regarded as intra-view negative samples, while embeddings of other nodes in the other view are regarded as inter-view negative samples. The contrastive loss is defined as:
\begin{equation}
    \ell\left(\boldsymbol{u_i}, \boldsymbol{u_i}^\prime\right)=\log \frac{e^{\theta\left(\boldsymbol{u_i}, \boldsymbol{u_i}^\prime\right) / \tau}}{\underbrace{e^{\theta\left(\boldsymbol{u_i}, \boldsymbol{u_i}^\prime\right) / \tau}}_\text{the positive pair}+\underbrace{\sum_{j=1}^N \mathrm{1}_{[j \neq i]} e^{\theta\left(\boldsymbol{u_i}, \boldsymbol{u_j}\right) / \tau}}_\text{intra-view negative pairs}+\underbrace{\sum_{j=1}^N \mathrm{1}_{[j \neq i]} e^{\theta\left(\boldsymbol{u_i}, \boldsymbol{u_j}^\prime\right) / \tau}}_\text{inter-view negative pairs}},
    \label{loss_ins}
\end{equation}
where $\mathrm{1}_{[j \neq i]} \in \left \{ 0, 1 \right \}$ is an indicator function that equals 1 iff $j \neq i$, $\theta\left(\cdot, \cdot\right)$ is the cosine similarity function, and $\tau$ is a temperature parameter. The loss for the other view is similarly defined, and the overall objective $\mathcal{L}_{ins}$ is the average of all instances:
\begin{equation}
    \mathcal{L}_{ins} = \frac{1}{2N} \sum_{i=1}^{N} \left [ \ell\left(\boldsymbol{u_i}, \boldsymbol{u_i}^\prime\right) + \ell\left(\boldsymbol{u_i}^\prime, \boldsymbol{u_i}\right) \right ].
\end{equation}

To enhance the scalability of our method for large-scale graphs, we employ the subsampling approach proposed by~\cite{SAGE}. Both the RE and MF methods, along with the loss function described in Equation~\ref{loss_ins}, are seamlessly adaptable to the sampled subgraphs.

\subsubsection{Feature-wise contrastive learning with token embeddings}\label{text_proto}

Instance-wise contrastive learning relies heavily on individual instances, which can cause transfer issues when transitioning to other datasets. Moreover, there is a significant gap between the obtained node representations and the semantic space of LLMs. To address these issues, we propose feature-wise contrastive learning with token embeddings.

Feature-wise contrastive loss breaks the independence between instances. For the feature matrix $\mathbf{U_\ast}$, we denote the columns in different views as $\boldsymbol{m_i} \in \mathbf{U_1^\top}$ and $\boldsymbol{n_i} \in \mathbf{U_2^\top}$. Here, $\boldsymbol{m_i}, \boldsymbol{n_i} \in \mathbb{R}^{N}$. The loss is denoted as $\mathcal{L}_{fea}$, and is calculated as:
\begin{equation}
    \mathcal{L}_{fea}=\frac{1}{F_U} \sum_{i=1}^{F_U}\log \frac{e^{\theta\left(\boldsymbol{m_i}, \boldsymbol{n_i}\right) / \tau}} {\sum_{j=1}^{F_U}\left[{e^{\theta\left(\boldsymbol{m_i}, \boldsymbol{m_j}\right) / \tau}}+{e^{\theta\left(\boldsymbol{m_i}, \boldsymbol{n_j}\right) / \tau}}\right]}.
    \label{loss_fea}
\end{equation}

To map node representations to the semantic space of LLMs, we use the principal components of the token embeddings of LLMs as coordinate axes. This approach ensures that the representations of similar instances are closely aligned in the textual embedding space. This helps alleviate the inconsistency in optimization objectives during graph self-supervised learning due to the gap between node representations and the text embedding space.

Specifically, we first use principal component analysis (PCA) to obtain the $P$ principal components, denoted as $\mathbf{C} \in \mathbb{R}^{P \times F_L}$, where $F_L$ is the dimension size of token embeddings of LLM. Then, we map node representations by:
\begin{equation}
    \widetilde{\mathbf{U}}_\ast = \mathbf{U}_\ast \times \mathbf{C}^\top.
\end{equation}
To map the node representations obtained from the GNN using principal components, we set the output dimension of the GNN to be equal to the token embeddings' dimension~(i.e., $F_U=F_L$). The columns of the mapped feature matrix $\widetilde{\mathbf{U}}_\ast$, denoted as $\widetilde{\boldsymbol{m}}_i$ and $\widetilde{\boldsymbol{n}}_i$, are fed into $\mathcal{L}_{fea}$. Therefore, the final contrastive loss for graph self-supervised learning is the average of Equation \ref{loss_ins} and Equation \ref{loss_fea}:
\begin{equation}
\mathcal{L}=\frac{1}{2} \left( \mathcal{L}_{ins} + \mathcal{L}_{fea} \right).
\label{loss_all}
\end{equation}

\textbf{\emph{Remark:}}  The introduction of feature-wise contrastive learning with token embeddings successfully addresses the semantic space discrepancy between graph node representations and LLM token embeddings. Our method enables the direct and simple use of graph structural and text information obtained by GNN in LLMs, thereby avoiding the significant generalization loss associated with complex modality alignment training during the fine-tuning process. Its role in fine-tuning will be further described in Sec.~\ref{graph2token} and validated by experiments. Additionally, the feature-wise contrastive method itself exhibits stronger generalization, allowing it to perform well on unseen instances~(or tasks) rather than relying on trained instances~(or tasks).

\subsection{Alignment tuning}

The development of LLMs has introduced a new paradigm for graph machine learning. However, existing research~\cite{canllm} indicates that LLMs alone cannot fully comprehend graph structures and their underlying information. To enable LLMs to more effectively capture information and improve their performance in cross-dataset and cross-task zero-shot learning, it is essential to design specific methods for LLMs to incorporate graph information suitably. To this end, we propose an alignment tuning method that includes specially designed instructions for various graph tasks at different levels, as well as a graph representation to graph token embeddings mechanism to integrate graph information.

\subsubsection{Instructions design}
\label{instruction_design}
The instruction we designed can be divided into two parts: one part provides graph information, and the other part describes the task. Here, we take a citation graph as an example, where nodes are papers, and relations are citations, to introduce the instruction.

\paragraph{Graph information provision} 
The graph information provision in the instructions for node, edge, and graph-level tasks is presented as follows: \textit{Given the representation of a paper/two papers/a paper set: $\mathsf{\langle graph \rangle}$, with the following information:\textbackslash nTitle: First Paper: $\mathsf{\left \{ title_1 \right \}}$ ...\textbackslash n}, where $\mathsf{\langle graph \rangle}$ is the placeholder for graph inputs (see Sect.~\ref{graph2token}), and $\mathsf{\left \{ title_1 \right \}}$ is the node text information.

Note that, different from most work which use LLM as a predictor, the instruction we designed uses only the title of a paper node, excluding more extensive textual information such as its abstract or description. In fact, reducing the amount of input text not only does not decrease the model's performance but actually improves it. \cite{canllm} confirmed through experiments that LLMs benefit from structural information only when the target node lacks sufficient phrases for reasonable predictions. Therefore, using only titles as text input can help LLMs extract more critical information from graph information. The complete instruction for the tasks of node classification and link prediction in citation networks is shown in Appendix~\ref{sec:instruction}.

\paragraph{Task description} 
To achieve cross-dataset capability, where the model can be trained on one graph dataset and then perform reasoning on any other dataset, the instruction is designed to include not only the task description itself but also the set of alternative answers. Using the node classification task on the Arxiv dataset (see Sect.~\ref{sect:setup}) as an example, the instruction is structured as follows: \textit{Which arXiv CS sub-category does this paper belong to? Please directly give the most likely answer from the following sub-categories: $\mathsf{\left \{ ans \right \}}$}, where $\mathsf{\left \{ ans \right \}}$ represents the set of alternative answers, which varies across datasets. Including alternative answers enables the model to learn the task of ``reasoning the answer from a given set according to the task'' rather than memorizing answers for a particular dataset, thus facilitating reasoning across datasets.


\subsubsection{Graph token embeddings}
\label{graph2token}

The token embeddings of graph mentioned previously, {\it i.e.}, $\mathsf{\langle graph \rangle}$, are crucial for incorporating graph information and enabling the model's generalization. We use a projector to map central node representations into $K$ graph token embeddings and replace $\mathsf{\langle graph \rangle}$ with these tokens. Kindly note that, we map the representations to fixed number of token embeddings regardless of the task type. For example, for node-level tasks, we map the central node representation to $K$ token embeddings; for edge-level tasks, we pool the representations of the two nodes of the target edge and then map this pooled representation to $K$ token embeddings; for graph-level tasks, similar approach can be applied. In this way, we unify the instruction of graph tasks at different levels. Thanks to the text-aligned contrastive learning, a linear projector is enough to capture the map relationship without tuning LLM:
\begin{equation}
    \mathbf{H}_{token} = f_\mathrm{Linear}\left(\boldsymbol{u_i}\right)
\end{equation}
where $\boldsymbol{u_i} \in \mathbf{U}$, $\mathbf{H}_{token} \in \mathbb{R}^{K \times F_L}$, $F_L$ is the dimension size of token embedding of LLM, and $f_\mathrm{Linear}(\cdot)$ is a linear layer.

\textbf{\emph{Remark:}} This approach offers three primary advantages: (\romannumeral1) When handling tasks at different levels, the changes to the instructions are minimal. This consistency facilitates the transfer of knowledge learned during training to unseen tasks in large language models (LLMs); (\romannumeral2) The fixed number of token embeddings can be seen as a conditional soft prompt. Unlike traditional soft prompts, learning at the instance level reduces the risk of overfitting to specific datasets or tasks, thereby enhancing generalization to unseen datasets and tasks; (\romannumeral3) Different from current work which intends to include the representations of all nodes in the subgraph, we only map the representations of the central node to tokens, since there has enough information carried by message passing of GNN. This method is more efficient, and it offers greater generalizability and practicality.

\subsubsection{Training and evaluation strategy}
\label{train_strategy}
To ensure compatibility and facilitate comparisons across various datasets, we map the node features into a consistent vector space. Specifically, we employ a pretrained BERT model \cite{devlin2018bert} to encode the raw text associated with each node, thereby generating the node features. We then pretrain the graph model using contrastive learning with the loss function defined in Equation \ref{loss_all} on a single dataset. After pretraining, the model parameters are fixed. We utilize the pretrained model to obtain node representations and follow the instructions in Section \ref{instruction_design} to train the linear projector on specific tasks within the same dataset. Finally, we evaluate the performance of our model on unseen datasets and tasks. Throughout all phases, the parameters of the language model remain fixed. We use GraphSAGE \cite{SAGE} as our graph encoder and Vicuna-7B-v1.5 \cite{vicuna2023} as the foundational language model.

\section{Experimental results}
\label{sect:exper}
In this section, comprehensive experiments are conducted to validate the effectiveness of TEA-GLM. These experiments aim to investigate the following research questions:
\begin{itemize}[leftmargin=28pt]
    \itemsep0em 
    \item[\textbf{RQ1:}] How effective is TEA-GLM in handling the cross-dataset zero-shot learning problem?
    \item[\textbf{RQ2:}] How well does TEA-GLM transfer knowledge when adapted to an unseen task and dataset in a zero-shot setting?
    \item[\textbf{RQ3:}] What is the contribution of the feature-wise contrastive learning and graph token embeddings to the zero-shot learning ability of TEA-GLM?
\end{itemize}

\subsection{Experimental setup}
\label{sect:setup}

\paragraph{Datasets}
We test TEA-GLM across eight widely used datasets spanning two distinct domains. Within the citation domain, we employ Arxiv~\cite{OGB}, Pubmed~\cite{Pubmed}, and an expanded version of Cora~\cite{Cora} with an increased range of classes and larger scale. In these datasets, each node represents an individual paper, with edges indicating citation relationships. In the e-commerce domain, we utilize datasets from the TAG benchmark~\cite{TAG}, including Children (Book-Children), History (Book-History), Computer (Ele-Computer), Photo (Ele-Photo), and Sports (Sports-Fitness). Here, nodes represent distinct products, while edges denote co-viewing or co-purchasing between two products. Appendix~\ref{sec:data_stats} presents the statistics for these datasets.

\paragraph{Baselines}
We conduct a comprehensive comparison of TEA-GLM with various categories of baseline methods: (\romannumeral1) Non-graph neural network approaches, such as MLP, which employs a Multilayer Perceptron for node representation; (\romannumeral2) Supervised methods, including GCN~\cite{GCN}, GraphSAGE~\cite{SAGE}, and GAT~\cite{GAT}; (\romannumeral3) Self-supervised methods like DGI~\cite{DGI}, which maximizes mutual information to learn node representations without relying on ground truth labels; (\romannumeral4) Graph knowledge distillation frameworks: GKD~\cite{GKD}, which distills knowledge from a teacher GNN trained on a complete graph to a student GNN operating on a smaller or sparser graph; GLNN~\cite{GLNN}, a method combining the advantages of graph neural networks and MLPs using knowledge distillation, aimed at reducing dependency on the inference graph; (\romannumeral5) Graph transformer networks, including NodeFormer~\cite{NodeFormer} and DIFFormer~\cite{DIFFormer}; (\romannumeral6) Large language models, such as Vicuna-7B-v1.5; (\romannumeral7) The latest models equipped with transfer and zero-shot capabilities, such as OFA~\cite{OFA}, GraphGPT~\cite{graphgpt}, and LLaGA~\cite{llaga}.

\paragraph{Implementation details}
\label{imple_detail}
For datasets within the citation domain, we follow the data split methodology outlined in GraphGPT~\cite{graphgpt}. For those within the e-commerce domain, we utilize scripts provided by the TAG benchmark~\cite{TAG} to generate data splits. To ensure comparability among different methods, identical data splits are applied to all models. To assess the performance of TEA-GLM, we employ three commonly adopted evaluation metrics: Accuracy and Macro F1 for node classification, and AUC (Area Under the Curve) for link prediction. To ensure result robustness, we conduct five experiments with random seed values ranging from 0 to 4 and report the mean and standard deviation of the results. Due to the limited number of pages, several experimental results, such as Macro F1 results of node classification~(Appendix~\ref{marco_f1}), legality rate of valid answers produced by the LLM~(Appendix~\ref{sec:legality}), and parameter sensitivity analysis~(Appendix~\ref{Sensitivity}), are reported in Appendix.

In the pre-training phase of the GNN, we set the GNN layers to 2. We use a batch size of 512 for 60 epochs and a learning rate of $2 \times 10^{-2}$. During the training of the linear projector, we configure a batch size of 2 per GPU for one epoch, with a learning rate of $1 \times 10^{-3}$. The Adam optimizer is employed for all approaches. For baseline models, we adjust hyperparameters and utilize the optimal settings. All experiments are conducted on 2 NVIDIA A100 GPUs with 80GB memory each, using CUDA version 11.7.

\begin{table}[t!]
    \centering
    \caption{Zero-shot accuracy on citation and e-commerce datasets (\textbf{bold} highlights the best result across all methods, while \underline{underline} highlights the second-best results)}
    \label{tab:nc_citation}
    \Large
    \resizebox{\textwidth}{!}{%
    \begin{tabular}{cc|cc|cccc}
    \toprule
    \multirow{2}{*}{\textbf{Model type}} & \multirow{2}{*}{\textbf{Model}} & \multicolumn{2}{c|}{\textbf{Citation}} & \multicolumn{4}{c}{\textbf{E-commerce}} \\
    \cmidrule{3-8}
    & & \textbf{Pubmed} & \textbf{Cora} & \textbf{Children} & \textbf{History} & \textbf{Photo} & \textbf{Sports} \\
    \midrule
    & MLP & 0.323$\pm$0.027 & 0.021$\pm$0.006 & 0.029$\pm$0.037 & 0.080$\pm$0.041 & 0.110$\pm$0.070 & 0.042$\pm$0.021 \\
    \midrule
    \multirow{9}{*}{\shortstack{GNN as\\ predictor}} 
    & GCN & 0.288$\pm$0.092 & 0.017$\pm$0.004 & 0.030$\pm$0.018 & 0.063$\pm$0.042 & 0.103$\pm$0.047 & 0.042$\pm$0.025 \\
    & GraphSAGE & 0.316$\pm$0.058 & 0.014$\pm$0.007 & 0.008$\pm$0.007 & 0.195$\pm$0.206 & 0.056$\pm$0.055 & 0.051$\pm$0.015 \\
    & GAT & 0.343$\pm$0.064 & 0.016$\pm$0.004 & 0.086$\pm$0.084 & 0.172$\pm$0.098 & 0.050$\pm$0.027 & 0.142$\pm$0.138 \\
    & DGI & 0.329$\pm$0.103 & 0.026$\pm$0.009 & 0.082$\pm$0.035 & 0.218$\pm$0.168 & 0.224$\pm$0.127 & 0.049$\pm$0.017 \\
    & GKD & 0.399$\pm$0.033 & 0.042$\pm$0.008 & 0.202$\pm$0.064 & 0.339$\pm$0.138 & 0.166$\pm$0.086 & 0.208$\pm$0.077 \\
    & GLNN & 0.390$\pm$0.011 & 0.031$\pm$0.006 & 0.187$\pm$0.012 & 0.283$\pm$0.021 & \underline{0.403$\pm$0.019} & 0.317$\pm$0.048 \\
    & NodeFormer & 0.308$\pm$0.093 & 0.016$\pm$0.007 & 0.048$\pm$0.028 & 0.168$\pm$0.127 & 0.073$\pm$0.015 & 0.165$\pm$0.057 \\
    & DIFFormer & 0.361$\pm$0.071 & 0.029$\pm$0.014 & 0.129$\pm$0.030 & 0.275$\pm$0.171 & 0.321$\pm$0.055 & 0.306$\pm$0.131 \\
    & OFA & 0.314$\pm$0.059 & 0.130$\pm$0.019 & 0.064$\pm$0.086 & 0.052$\pm$0.049 & 0.340$\pm$0.026 & 0.101$\pm$0.071 \\
    \midrule
    \multirow{5}{*}{\shortstack{LLM as\\ predictor}} 
    & Vicuna-7B-v1.5 & 0.719$\pm$0.010 & 0.156$\pm$0.001 & \underline{0.270$\pm$0.001} & \underline{0.363$\pm$0.001} & 0.378$\pm$0.004 & \underline{0.370$\pm$0.001} \\
    & Vicuna-7B-SPT & 0.768$\pm$0.036 & 0.168$\pm$0.018 & 0.227$\pm$0.015 & 0.281$\pm$0.088 & 0.350$\pm$0.061 & 0.230$\pm$0.018 \\
    & GraphGPT-std & 0.701 & 0.126 & - & - & - & - \\
    & GraphGPT-cot & 0.521 & \underline{0.181} & - & - & - & - \\
    & LLaGA & \underline{0.793$\pm$0.036} & 0.168$\pm$0.032 & 0.199$\pm$0.007 & 0.146$\pm$0.067 & 0.276$\pm$0.069 & 0.352$\pm$0.033 \\
    & \textbf{TEA-GLM} & \textbf{0.848$\pm$0.010} & \textbf{0.202$\pm$0.014} & \textbf{0.271$\pm$0.010} & \textbf{0.528$\pm$0.058} & \textbf{0.497$\pm$0.027} & \textbf{0.404$\pm$0.010} \\
    \bottomrule
    \end{tabular}
    }
\end{table}

\subsection{Cross-dataset zero-shot ability (RQ1)}
\label{sect:cross-dataset}

We train all methods on the Arxiv and Computer, respectively, followed by an evaluation of their zero-shot performance on datasets from the same domain. Zero-shot learning presents challenges for GNN-based models, particularly regarding variations in the number of classes across different datasets. To address this, we adopt the setting outlined in GraphGPT~\cite{graphgpt}. For each target dataset, we utilize the GNN backbone trained on the source dataset along with a classifier trained with target data, typically a linear layer. Due to the considerable time cost associated with training and evaluating GraphGPT on e-commerce datasets, we only report its performance on citation datasets as provided in their paper. ``-std'' and ``-cot'' denote the use of the standard procedure of dual-stage graph instruction tuning and COT instruction datasets generated by LLM, respectively. To demonstrate the difference between our work and Soft Prompt Tuning, we fine-tuned vicuna-7b-v1.5 using Soft Prompt and reported the results. The Accuracy results are presented in Table~\ref{tab:nc_citation}. As mentioned earlier, we report the Macro F1 results in Appendix~\ref{marco_f1} and report results on two training datasets in Appendix~\ref{supervised}.

The results clearly demonstrate that TEA-GLM outperforms all state-of-the-art (SOTA) models, resulting in significant improvements. Comparative analysis with baseline models across all datasets highlights the robust generalization capability of TEA-GLM. Models utilizing GNN as a predictor face challenges in achieving cross-dataset transferability with traditional supervised and self-supervised learning methods. Even recently developed robust GNN-based models, such as NodeFormer, DIFFormer, and GKD, encounter similar issues. In the case of OFA, a recent framework for cross-domain learning, strong transferability is observed between topic-related datasets such as Arxiv and Cora (both related to computer science). Nevertheless, its generalization performance notably decreases on datasets with lower topic relevance, such as those in the e-commerce domain.

LLM-based solutions, such as Vicuna-7B, demonstrate consistent performance across various datasets. Nevertheless, their predictive capabilities are confined to text information alone. Vicuna-7B-SPT also fails to achieve transferability on e-commerce datasets, indicating that soft prompt tuning alone is insufficient when relying solely on node texts. This suggests that graph tokens indeed contain transferable graph information, enabling the LLM to make more accurate predictions. In contrast, GNN-LLM-combined solutions that use LLM as a predictor demonstrate generalization ability but often face limitations. For instance, GraphGPT tends to underperform compared to Vicuna-7B, due to the lack of a graph foundation model. Instead of relying on a graph foundation model, LLaGA directly maps node representations without GNN and can generalize on citation datasets. However, it demonstrates limited generalization capability across e-commerce datasets, which are more challenging due to highly irrelevant topics. TEA-GLM, on the other hand, utilizes principal components of token embeddings of LLMs to constrain representations learned by GNN, helping the graph representations well transfer to other datasets. Experimental results validate the superior generalization capabilities of TEA-GLM, achieved with less textual data and fewer parameters.

\subsection{Cross-task zero-shot ability (RQ2)}
\label{sect:cross-task}
We employ models trained on node classification tasks directly for link prediction tasks without any fine-tuning. We omit the comparison with models utilizing GNN as a predictor, as conducting cross-task evaluation of these models without fine-tuning poses a significant challenge, given that different tasks typically correspond to different task heads. Here, we contrast TEA-GLM with OFA, which similarly enables cross-task testing without the need for fine-tuning. Additionally, we compare TEA-GLM with Vicuna-7B and methods that utilize LLM as a predictor, such as GraphGPT and LLaGA. For GraphGPT, we utilize the checkpoint released by the author trained on Arxiv and report the results on citation datasets. The results are reported in Table~\ref{cross_task}.

In the case of OFA, although this framework facilitates cross-domain and cross-task learning, it exhibits negative transfer when lacking task-relevant data, particularly on unseen tasks. Benefiting from the generalization capability of large language models, both the fine-tuned and non-fine-tuned versions of Vicuna do not experience negative transfer. However, due to the absence of graph information, its predictions often appear random. Conversely, GraphGPT shows transferability with familiar datasets, yet its performance declines when dealing with unseen datasets~(Pubmed and Cora). Due to the absence of GNN for filtering and aggregating graph information, LLaGA demonstrates unstable performance. While it exhibits cross-task transferability on citation datasets, its performance is poor on most e-commerce datasets. In contrast, TEA-GLM consistently outperforms all baseline methods on both unseen datasets and tasks, except for the results on Sports, indicating the stronger generalization ability of TEA-GLM.

\begin{table}[t!]
    \centering
    \caption{AUC of link prediction~(Cross-task)}
    \label{cross_task}
    \resizebox{0.9\textwidth}{!}{%
    \begin{tabular}{c|ccc|ccccc}
    \toprule
    \multirow{2}{*}{\textbf{Model}} & \multicolumn{3}{c|}{\textbf{Citation}} & \multicolumn{5}{c}{\textbf{E-commerce}} \\
    \cmidrule{2-9}
    & \textbf{Arxiv} & \textbf{Pubmed} & \textbf{Cora} & \textbf{Children} & \textbf{History} & \textbf{Computer} & \textbf{Photo} & \textbf{Sports} \\
    \midrule
    OFA & 0.469 & 0.481 & 0.492 & 0.484 & 0.431 & 0.461 & 0.459 & 0.517 \\
    Vicuna-7B-v1.5 & 0.513 & 0.543 & 0.527 & 0.500 & 0.515 & 0.502 & \underline{0.501} & 0.502 \\
    Vicuna-7B-SPT & 0.537 & 0.535 & \underline{0.565} & \underline{0.544} & \underline{0.543} & \underline{0.509} & \underline{0.501} & 0.508 \\
    GraphGPT-std & \underline{0.649} & 0.501 & 0.520 & - & - & - & - & - \\
    LLaGA & 0.570 & \underline{0.569} & 0.537 & 0.422 & 0.449 & 0.479 & 0.478 & \textbf{0.597} \\
    \midrule
    \textbf{TEA-GLM} & \textbf{0.657} & \textbf{0.689} & \textbf{0.586} & \textbf{0.571} & \textbf{0.579} & \textbf{0.554} & \textbf{0.545} & \underline{0.553} \\
    \bottomrule
    \end{tabular}
    }
\end{table}

\subsection{Ablation study (RQ3)}
\label{sect:ablation}

\begin{figure}[t!]
    \centering
    \subfigure[Accuracy of node classification]{
    \label{ablation_nc}
        \includegraphics[scale=0.242]{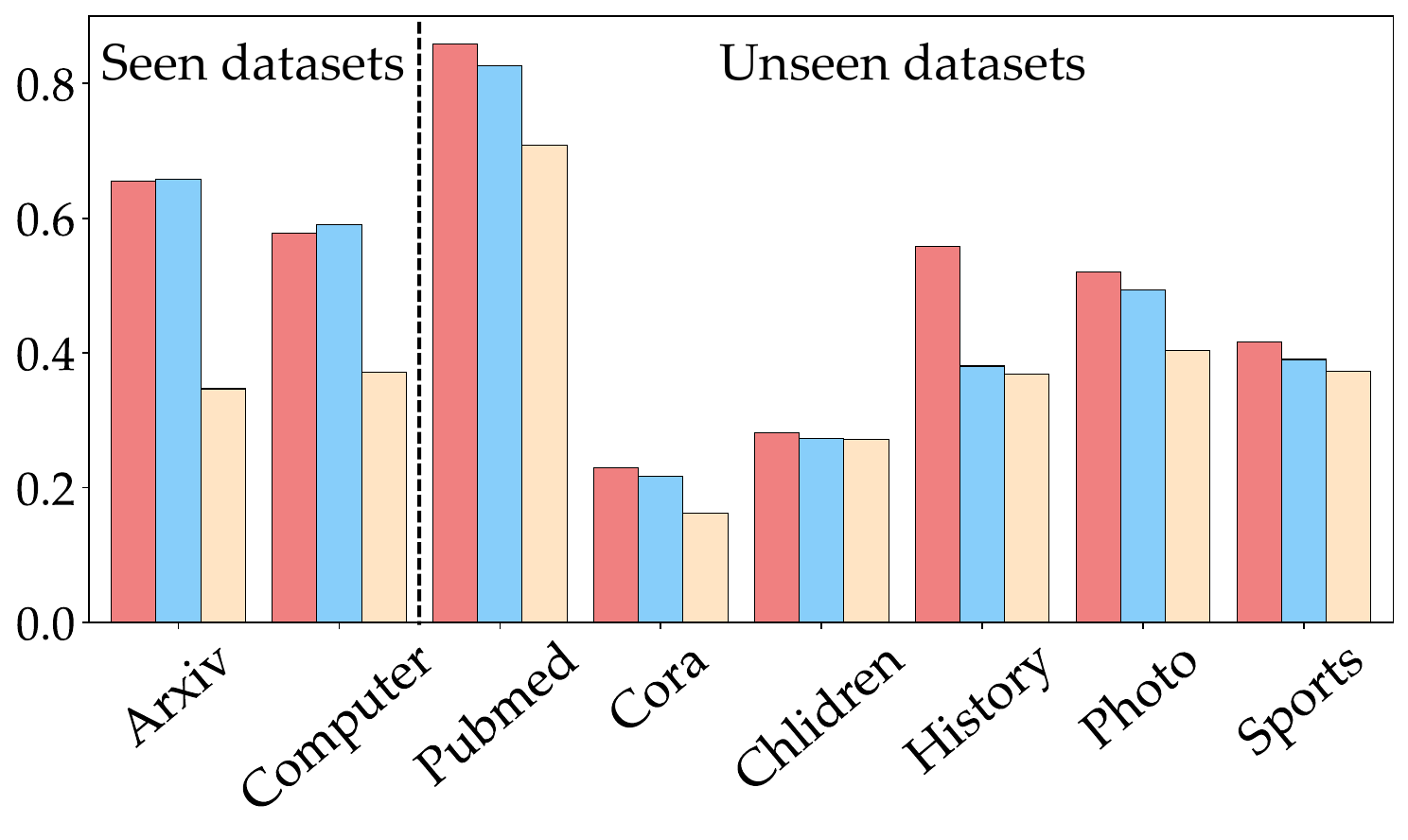}
    }
    \subfigure[AUC of link prediction]{
    \label{ablation_lp}
        \includegraphics[scale=0.242]{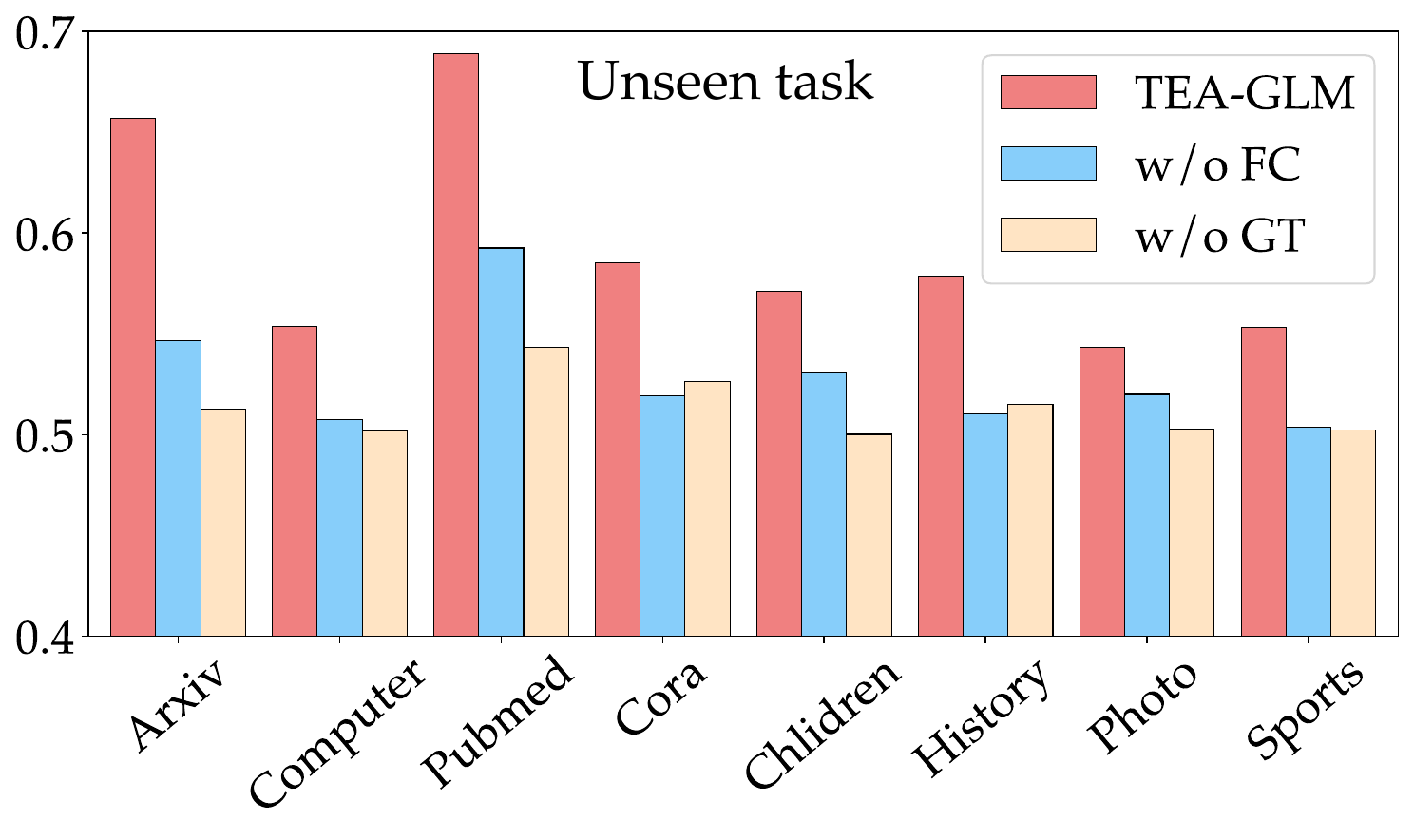}
    }
    \caption{Ablation study results~(``Seen datasets'' are used to train the GNN and linear projector, while ``unseen datasets'' are not. ``Unseen task'' means the model wasn't trained for link prediction.)}
    \label{ablation}
\end{figure}

We conduct an ablation study to discuss two key components of our model: feature-wise contrastive learning and graph token embeddings. Here, we directly remove these two components from our model and then test the model's performance on cross-dataset and cross-task evaluations. The results are shown in Figure~\ref{ablation}. ``w/o FC'' means that we pretrain the GNN without feature-wise contrastive learning, while ``w/o GT'' means predicting without graph token embeddings.

Without graph token embeddings, large language models lack crucial information from the graph, leading to a significant decline in performance on both node-level and edge-level tasks. GNNs pre-trained with feature-wise contrastive learning can obtain node representations aligned with the text space, enabling cross-dataset and cross-task generalization through a simple linear layer. When the feature-wise constraint for pre-training is absent, the model's performance on the seen datasets (Arxiv and Computer) for the training task improves slightly. However, its performance on unseen datasets declines. Although it remains relatively stable when handling tasks of the same category, its performance decreases notably when dealing with unseen tasks (link prediction). These results indicate that alignment between graph representation and LLM's token embeddings via feature-wise contrastive learning is important for cross-task zero-shot transfer.

\section{Related work}
\label{sect:related_work}

\subsection{Graph neural networks}
\label{sect:GNN}
In the field of graph machine learning, Graph Neural Networks~(GNNs) have garnered significant attention~\cite{gnn_work_1, gnn_work_2, gnn_work_3, gnn_work_4, gnn_work_5, gnn_work_6, gnn_work_7, gnn_work_8}. The primary strategy of most GNNs is to capture underlying message-passing patterns for graph representation. Several effective neural network architectures have been proposed, such as Graph Attention Network~(GAT)~\cite{GAT}, Graph Convolution Network~(GCN)~\cite{GCN}, and GraphSAGE~\cite{SAGE}. Recently, there has been a surge of interest in exploring transformer-based encoders for graph machine learning~\cite{gtf1, gtf2, NodeFormer, DIFFormer}. However, a notable limitation of GNNs is their generalization capability. Typically, GNNs are trained on specific tasks within particular datasets, and when faced with new datasets or tasks, they often struggle to consistently perform well across different datasets or downstream tasks~\cite{ParetoGNN}.

\subsection{Self-supervised learning and prompt-tuning for GNNs}
\label{sect:self-supervised}
To alleviate the demand for labeled data and enhance the robustness of graph models, self-supervised learning is commonly employed in GNN training~\cite{simgrace, zhu2020deep, mvgrl}. Methods like Deep Graph Infomax~(DGI)~\cite{DGI} utilize mutual information maximization for pre-training. Other approaches, such as GraphCL~\cite{GraphCL}, GCA~\cite{GCA}, GCC~\cite{GCC}, and JOAO~\cite{JOAO}, learn node representations by contrasting positive and negative samples. GraphMAE~\cite{graphmae, graphmae2}, on the other hand, learns representations by generating samples that resemble the original graph structure. However, these methods typically require fine-tuning the task-specific heads for downstream applications.

Various methods have explored the use of prompt techniques to enhance the generalization of GNNs. To address the inconsistency between pre-training and downstream task objectives, GraphPrompt~\cite{graphprompt} proposes a unified task template applicable to both stages. Additionally, ProG~\cite{allinone} reformulates various task types into a unified graph-level representation and employs meta-learning techniques to enhance multi-task learning capabilities. However, whether through self-supervised learning or graph prompt methods, fine-tuning is often necessary when handling new datasets. Moreover, when confronted with datasets containing varying numbers of categories, retraining of task heads is required to achieve optimal performance.

\subsection{Large language models for graphs}
\label{LLM4G}
With the rapid advancement of Large Language Models~(LLMs) and their remarkable generalization capabilities, leveraging LLMs to address transferability issues in graph machine learning has garnered significant attention~\cite{gpt4graph, unigraph}. Some methods represent graph structure information as text input to LLMs~\cite{graph2text, nlgraph, llm2graph}; however, this approach often leads to suboptimal solutions~\cite{canllm}. Another paradigm involves using LLMs as enhancers~\cite{natureisneed, llmgnn1, llmgnn2, labelfree, OFA}, where they generate data or node text representations. Despite this, since GNNs are ultimately used for prediction, this approach significantly limits the model's transferability. Recently, considerable efforts have been made to utilize LLMs as predictors. For instance, GraphGPT~\cite{graphgpt} attempts to align LLMs with pre-trained Graph Transformer encoders through two-stage fine-tuning. However, the fine-tuning, conducted on specific datasets, might weaken the method's transferability. In light of this, LLaGA~\cite{llaga} introduced a novel encoding method that directly translates graph data into sequences compatible with LLMs. However, this approach may compromise performance due to the lack of GNN filtering and aggregation of graph information. Inspired by these challenges, we propose a pre-training strategy that enhances GNN transferability by aligning its representations with the token embeddings of LLMs, resulting in improved performance in zero-shot tasks. Notably, similar to our method, TEST \cite{sun2023test} aligns time series representations with several selected LLM token embeddings. However, our approach differs in that we project graph representations into a feature space defined by the principal components of LLM token embeddings. This enables the LLM to function as a zero-shot learner for graph machine learning tasks, rather than just enhancing performance on specific, seen tasks.

\section{Limitations}
\label{sect:limitations}

While our TEA-GLM framework demonstrates considerable promise in enhancing zero-shot learning for graph-based tasks, it does have some limitations. Although the framework we designed can be easily applied to graph-level tasks, we have not yet explored the model's performance through specific experiments. This will be addressed in our future work.

\section{Conclusion}
\label{sect:conclusion}
This paper introduces TEA-GLM, a framework that enhances zero-shot learning in graph machine learning by aligning GNN representations with LLM token embeddings. TEA-GLM uses a linear projector to map graph representations into graph token embeddings and incorporates a unified instruction design to handle various graph tasks at different levels. This approach enables consistent performance across various datasets and tasks without task-specific fine-tuning. Extensive experiments show that TEA-GLM outperforms state-of-the-art methods in accuracy and generalization, demonstrating its effectiveness and efficiency in zero-shot learning for graph tasks.

\section{Acknowledgement}
\label{sect:acknowledgement}
This work was supported by the National Key R\&D Program of China (2023YFC3304700). The work of Yuan Zuo was supported by the National
Natural Science Foundation of China (NSFC) under Grant
71901012. Dr. Junjie Wu’s work was partially supported by the National Natural Science Foundation of China (72242101, 72031001) and Outstanding Young Scientist Program of Beijing Universities (JWZQ20240201002).

\bibliography{neurips_2024}

\begin{thebibliography}{53}
\providecommand{\natexlab}[1]{#1}
\providecommand{\url}[1]{\texttt{#1}}
\expandafter\ifx\csname urlstyle\endcsname\relax
  \providecommand{\doi}[1]{doi: #1}\else
  \providecommand{\doi}{doi: \begingroup \urlstyle{rm}\Url}\fi

\bibitem[Chen et~al.(2018)Chen, Ma, and Xiao]{gnn_work_8}
Jie Chen, Tengfei Ma, and Cao Xiao.
\newblock Fast{GCN}: Fast learning with graph convolutional networks via importance sampling.
\newblock In \emph{International Conference on Learning Representations}, 2018.
\newblock URL \url{https://openreview.net/forum?id=rytstxWAW}.

\bibitem[Chen et~al.(2024{\natexlab{a}})Chen, Zhao, Jaiswal, Shah, and Wang]{llaga}
Runjin Chen, Tong Zhao, Ajay Jaiswal, Neil Shah, and Zhangyang Wang.
\newblock Llaga: Large language and graph assistant.
\newblock In \emph{ICML}, 2024{\natexlab{a}}.

\bibitem[Chen et~al.(2023)Chen, Mao, Li, Jin, Wen, Wei, Wang, Yin, Fan, Liu, and Tang]{graph2text}
Zhikai Chen, Haitao Mao, Hang Li, Wei Jin, Hongzhi Wen, Xiaochi Wei, Shuaiqiang Wang, Dawei Yin, Wenqi Fan, Hui Liu, and Jiliang Tang.
\newblock Exploring the potential of large language models ({LLM}s) in learning on graph.
\newblock In \emph{NeurIPS 2023 Workshop: New Frontiers in Graph Learning}, 2023.
\newblock URL \url{https://openreview.net/forum?id=ScNNo7v4t0}.

\bibitem[Chen et~al.(2024{\natexlab{b}})Chen, Mao, Wen, Han, Jin, Zhang, Liu, and Tang]{labelfree}
Zhikai Chen, Haitao Mao, Hongzhi Wen, Haoyu Han, Wei Jin, Haiyang Zhang, Hui Liu, and Jiliang Tang.
\newblock Label-free node classification on graphs with large language models ({LLM}s).
\newblock In \emph{The Twelfth International Conference on Learning Representations}, 2024{\natexlab{b}}.
\newblock URL \url{https://openreview.net/forum?id=hESD2NJFg8}.

\bibitem[Cheng et~al.(2023)Cheng, Li, Li, and Tsung]{gnn_work_1}
Jiashun Cheng, Man Li, Jia Li, and Fugee Tsung.
\newblock Wiener graph deconvolutional network improves graph self-supervised learning.
\newblock In \emph{AAAI}, 2023.
\newblock URL \url{https://doi.org/10.1609/aaai.v37i6.25870}.

\bibitem[Chiang et~al.(2019)Chiang, Liu, Si, Li, Bengio, and Hsieh]{gnn_work_6}
Wei-Lin Chiang, Xuanqing Liu, Si~Si, Yang Li, Samy Bengio, and Cho-Jui Hsieh.
\newblock Cluster-gcn: An efficient algorithm for training deep and large graph convolutional networks.
\newblock In \emph{Proceedings of the 25th ACM SIGKDD international conference on knowledge discovery \& data mining}, pages 257--266, 2019.

\bibitem[Chiang et~al.(2023)Chiang, Li, Lin, Sheng, Wu, Zhang, Zheng, Zhuang, Zhuang, Gonzalez, Stoica, and Xing]{vicuna2023}
Wei-Lin Chiang, Zhuohan Li, Zi~Lin, Ying Sheng, Zhanghao Wu, Hao Zhang, Lianmin Zheng, Siyuan Zhuang, Yonghao Zhuang, Joseph~E. Gonzalez, Ion Stoica, and Eric~P. Xing.
\newblock Vicuna: An open-source chatbot impressing gpt-4 with 90\%* chatgpt quality, 2023.
\newblock URL \url{https://lmsys.org/blog/2023-03-30-vicuna/}.

\bibitem[Devlin et~al.(2019)Devlin, Chang, Lee, and Toutanova]{devlin2018bert}
Jacob Devlin, Ming-Wei Chang, Kenton Lee, and Kristina Toutanova.
\newblock {BERT}: Pre-training of deep bidirectional transformers for language understanding.
\newblock In \emph{Proceedings of the 2019 Conference of the North {A}merican Chapter of the Association for Computational Linguistics: Human Language Technologies, Volume 1 (Long and Short Papers)}, pages 4171--4186, 2019.
\newblock URL \url{https://aclanthology.org/N19-1423}.

\bibitem[Gao et~al.(2018)Gao, Wang, and Ji]{gnn_work_5}
Hongyang Gao, Zhengyang Wang, and Shuiwang Ji.
\newblock Large-scale learnable graph convolutional networks.
\newblock In \emph{Proceedings of the 24th ACM SIGKDD International Conference on Knowledge Discovery \& Data Mining}, page 1416–1424, 2018.
\newblock URL \url{https://doi.org/10.1145/3219819.3219947}.

\bibitem[Guo et~al.(2023)Guo, Du, and Liu]{gpt4graph}
Jiayan Guo, Lun Du, and Hengyu Liu.
\newblock Gpt4graph: Can large language models understand graph structured data ? an empirical evaluation and benchmarking.
\newblock \emph{ArXiv}, abs/2305.15066, 2023.
\newblock URL \url{https://api.semanticscholar.org/CorpusID:258865990}.

\bibitem[Hamilton et~al.(2017)Hamilton, Ying, and Leskovec]{SAGE}
Will Hamilton, Zhitao Ying, and Jure Leskovec.
\newblock Inductive representation learning on large graphs.
\newblock In \emph{Advances in Neural Information Processing Systems}, 2017.
\newblock URL \url{https://proceedings.neurips.cc/paper_files/paper/2017/file/5dd9db5e033da9c6fb5ba83c7a7ebea9-Paper.pdf}.

\bibitem[Hassani and Khasahmadi(2020)]{mvgrl}
Kaveh Hassani and Amir~Hosein Khasahmadi.
\newblock Contrastive multi-view representation learning on graphs.
\newblock In \emph{Proceedings of the 37th International Conference on Machine Learning}, 2020.

\bibitem[He et~al.(2024)He, Bresson, Laurent, Perold, LeCun, and Hooi]{Pubmed}
Xiaoxin He, Xavier Bresson, Thomas Laurent, Adam Perold, Yann LeCun, and Bryan Hooi.
\newblock Harnessing explanations: {LLM}-to-{LM} interpreter for enhanced text-attributed graph representation learning.
\newblock In \emph{The Twelfth International Conference on Learning Representations}, 2024.
\newblock URL \url{https://openreview.net/forum?id=RXFVcynVe1}.

\bibitem[He and Hooi(2024)]{unigraph}
Yufei He and Bryan Hooi.
\newblock Unigraph: Learning a cross-domain graph foundation model from natural language.
\newblock \emph{arXiv preprint arXiv:2402.13630}, 2024.

\bibitem[Hou et~al.(2022)Hou, Liu, Cen, Dong, Yang, Wang, and Tang]{graphmae}
Zhenyu Hou, Xiao Liu, Yukuo Cen, Yuxiao Dong, Hongxia Yang, Chunjie Wang, and Jie Tang.
\newblock Graphmae: Self-supervised masked graph autoencoders.
\newblock In \emph{Proceedings of the 28th ACM SIGKDD Conference on Knowledge Discovery and Data Mining}, page 594–604, 2022.
\newblock URL \url{https://doi.org/10.1145/3534678.3539321}.

\bibitem[Hou et~al.(2023)Hou, He, Cen, Liu, Dong, Kharlamov, and Tang]{graphmae2}
Zhenyu Hou, Yufei He, Yukuo Cen, Xiao Liu, Yuxiao Dong, Evgeny Kharlamov, and Jie Tang.
\newblock Graphmae2: A decoding-enhanced masked self-supervised graph learner.
\newblock In \emph{Proceedings of the ACM Web Conference 2023}, page 737–746, 2023.
\newblock URL \url{https://doi.org/10.1145/3543507.3583379}.

\bibitem[Hu et~al.(2020)Hu, Fey, Zitnik, Dong, Ren, Liu, Catasta, and Leskovec]{OGB}
Weihua Hu, Matthias Fey, Marinka Zitnik, Yuxiao Dong, Hongyu Ren, Bowen Liu, Michele Catasta, and Jure Leskovec.
\newblock Open graph benchmark: Datasets for machine learning on graphs.
\newblock In \emph{Advances in Neural Information Processing Systems}, pages 22118--22133, 2020.
\newblock URL \url{https://proceedings.neurips.cc/paper_files/paper/2020/file/fb60d411a5c5b72b2e7d3527cfc84fd0-Paper.pdf}.

\bibitem[Huang et~al.(2023)Huang, Zhang, Mei, and Ma]{canllm}
Jin Huang, Xingjian Zhang, Qiaozhu Mei, and Jiaqi Ma.
\newblock Can llms effectively leverage graph structural information: when and why.
\newblock \emph{arXiv preprint arXiv:2309.16595}, 2023.

\bibitem[Ju et~al.(2023)Ju, Zhao, Wen, Yu, Shah, Ye, and Zhang]{ParetoGNN}
Mingxuan Ju, Tong Zhao, Qianlong Wen, Wenhao Yu, Neil Shah, Yanfang Ye, and Chuxu Zhang.
\newblock Multi-task self-supervised graph neural networks enable stronger task generalization.
\newblock In \emph{The Eleventh International Conference on Learning Representations}, 2023.
\newblock URL \url{https://openreview.net/forum?id=1tHAZRqftM}.

\bibitem[Kipf and Welling(2017{\natexlab{a}})]{GCN}
Thomas~N. Kipf and Max Welling.
\newblock Semi-supervised classification with graph convolutional networks.
\newblock In \emph{International Conference on Learning Representations}, 2017{\natexlab{a}}.
\newblock URL \url{https://openreview.net/forum?id=SJU4ayYgl}.

\bibitem[Kipf and Welling(2017{\natexlab{b}})]{cora_old}
Thomas~N. Kipf and Max Welling.
\newblock Semi-supervised classification with graph convolutional networks.
\newblock In \emph{International Conference on Learning Representations}, 2017{\natexlab{b}}.
\newblock URL \url{https://openreview.net/forum?id=SJU4ayYgl}.

\bibitem[Li et~al.(2019)Li, Han, Cheng, Su, Wang, Zhang, and Pan]{gnn_work_2}
Jia Li, Zhichao Han, Hong Cheng, Jiao Su, Pengyun Wang, Jianfeng Zhang, and Lujia Pan.
\newblock Predicting path failure in time-evolving graphs.
\newblock In \emph{Proceedings of the 25th ACM SIGKDD International Conference on Knowledge Discovery \& Data Mining}, page 1279–1289, 2019.
\newblock URL \url{https://doi.org/10.1145/3292500.3330847}.

\bibitem[Liu and Wu(2023)]{llm2graph}
Chang Liu and Bo~Wu.
\newblock Evaluating large language models on graphs: Performance insights and comparative analysis.
\newblock \emph{arXiv preprint arXiv:2308.11224}, 2023.

\bibitem[Liu et~al.(2024)Liu, Feng, Kong, Liang, Tao, Chen, and Zhang]{OFA}
Hao Liu, Jiarui Feng, Lecheng Kong, Ningyue Liang, Dacheng Tao, Yixin Chen, and Muhan Zhang.
\newblock One for all: Towards training one graph model for all classification tasks.
\newblock In \emph{The Twelfth International Conference on Learning Representations}, 2024.
\newblock URL \url{https://openreview.net/forum?id=4IT2pgc9v6}.

\bibitem[Liu et~al.(2023)Liu, Yu, Fang, and Zhang]{graphprompt}
Zemin Liu, Xingtong Yu, Yuan Fang, and Xinming Zhang.
\newblock Graphprompt: Unifying pre-training and downstream tasks for graph neural networks.
\newblock In \emph{Proceedings of the ACM Web Conference 2023}, page 417–428, 2023.
\newblock URL \url{https://doi.org/10.1145/3543507.3583386}.

\bibitem[Qiu et~al.(2020)Qiu, Chen, Dong, Zhang, Yang, Ding, Wang, and Tang]{GCC}
Jiezhong Qiu, Qibin Chen, Yuxiao Dong, Jing Zhang, Hongxia Yang, Ming Ding, Kuansan Wang, and Jie Tang.
\newblock Gcc: Graph contrastive coding for graph neural network pre-training.
\newblock In \emph{Proceedings of the 26th ACM SIGKDD International Conference on Knowledge Discovery \& Data Mining}, page 1150–1160, 2020.
\newblock URL \url{https://doi.org/10.1145/3394486.3403168}.

\bibitem[Sun et~al.(2024)Sun, Li, Li, and Hong]{sun2023test}
Chenxi Sun, Hongyan Li, Yaliang Li, and Shenda Hong.
\newblock {TEST}: Text prototype aligned embedding to activate {LLM}'s ability for time series.
\newblock In \emph{The Twelfth International Conference on Learning Representations}, 2024.
\newblock URL \url{https://openreview.net/forum?id=Tuh4nZVb0g}.

\bibitem[Sun et~al.(2021)Sun, Yin, Liu, Chen, Meng, Han, and Cao]{gnn_work_3}
Xiangguo Sun, Hongzhi Yin, Bo~Liu, Hongxu Chen, Qing Meng, Wang Han, and Jiuxin Cao.
\newblock Multi-level hyperedge distillation for social linking prediction on sparsely observed networks.
\newblock In \emph{Proceedings of the Web Conference 2021}, page 2934–2945, 2021.
\newblock URL \url{https://doi.org/10.1145/3442381.3449912}.

\bibitem[Sun et~al.(2023)Sun, Cheng, Li, Liu, and Guan]{allinone}
Xiangguo Sun, Hong Cheng, Jia Li, Bo~Liu, and Jihong Guan.
\newblock All in one: Multi-task prompting for graph neural networks.
\newblock In \emph{Proceedings of the 29th ACM SIGKDD Conference on Knowledge Discovery and Data Mining}, page 2120–2131, 2023.
\newblock URL \url{https://doi.org/10.1145/3580305.3599256}.

\bibitem[Tang et~al.(2023)Tang, Yang, Wei, Shi, Su, Cheng, Yin, and Huang]{graphgpt}
Jiabin Tang, Yuhao Yang, Wei Wei, Lei Shi, Lixin Su, Suqi Cheng, Dawei Yin, and Chao Huang.
\newblock Graphgpt: Graph instruction tuning for large language models.
\newblock \emph{arXiv preprint arXiv:2310.13023}, 2023.

\bibitem[Veličković et~al.(2018)Veličković, Cucurull, Casanova, Romero, Liò, and Bengio]{GAT}
Petar Veličković, Guillem Cucurull, Arantxa Casanova, Adriana Romero, Pietro Liò, and Yoshua Bengio.
\newblock Graph attention networks.
\newblock In \emph{International Conference on Learning Representations}, 2018.
\newblock URL \url{https://openreview.net/forum?id=rJXMpikCZ}.

\bibitem[Veličković et~al.(2019)Veličković, Fedus, Hamilton, Liò, Bengio, and Hjelm]{DGI}
Petar Veličković, William Fedus, William~L. Hamilton, Pietro Liò, Yoshua Bengio, and R~Devon Hjelm.
\newblock Deep graph infomax.
\newblock In \emph{International Conference on Learning Representations}, 2019.
\newblock URL \url{https://openreview.net/forum?id=rklz9iAcKQ}.

\bibitem[Wang et~al.(2023)Wang, Feng, He, Tan, Han, and Tsvetkov]{nlgraph}
Heng Wang, Shangbin Feng, Tianxing He, Zhaoxuan Tan, Xiaochuang Han, and Yulia Tsvetkov.
\newblock Can language models solve graph problems in natural language?
\newblock In \emph{Thirty-seventh Conference on Neural Information Processing Systems}, 2023.
\newblock URL \url{https://openreview.net/forum?id=UDqHhbqYJV}.

\bibitem[Wei et~al.(2022)Wei, Bosma, Zhao, Guu, Yu, Lester, Du, Dai, and Le]{finetunedlmforzs}
Jason Wei, Maarten Bosma, Vincent Zhao, Kelvin Guu, Adams~Wei Yu, Brian Lester, Nan Du, Andrew~M. Dai, and Quoc~V Le.
\newblock Finetuned language models are zero-shot learners.
\newblock In \emph{International Conference on Learning Representations}, 2022.
\newblock URL \url{https://openreview.net/forum?id=gEZrGCozdqR}.

\bibitem[Wen and Fang(2023)]{Cora}
Zhihao Wen and Yuan Fang.
\newblock Augmenting low-resource text classification with graph-grounded pre-training and prompting.
\newblock In \emph{Proceedings of the 46th International ACM SIGIR Conference on Research and Development in Information Retrieval}, page 506–516, 2023.
\newblock \doi{10.1145/3539618.3591641}.
\newblock URL \url{https://doi.org/10.1145/3539618.3591641}.

\bibitem[Wu et~al.(2022)Wu, Zhao, Li, Wipf, and Yan]{NodeFormer}
Qitian Wu, Wentao Zhao, Zenan Li, David Wipf, and Junchi Yan.
\newblock Nodeformer: A scalable graph structure learning transformer for node classification.
\newblock In Alice~H. Oh, Alekh Agarwal, Danielle Belgrave, and Kyunghyun Cho, editors, \emph{Advances in Neural Information Processing Systems}, 2022.
\newblock URL \url{https://openreview.net/forum?id=sMezXGG5So}.

\bibitem[Wu et~al.(2023)Wu, Yang, Zhao, He, Wipf, and Yan]{DIFFormer}
Qitian Wu, Chenxiao Yang, Wentao Zhao, Yixuan He, David Wipf, and Junchi Yan.
\newblock {DIFF}ormer: Scalable (graph) transformers induced by energy constrained diffusion.
\newblock In \emph{The Eleventh International Conference on Learning Representations}, 2023.
\newblock URL \url{https://openreview.net/forum?id=j6zUzrapY3L}.

\bibitem[Xia et~al.(2022)Xia, Wu, Chen, Hu, and Li]{simgrace}
Jun Xia, Lirong Wu, Jintao Chen, Bozhen Hu, and Stan~Z. Li.
\newblock Simgrace: A simple framework for graph contrastive learning without data augmentation.
\newblock In \emph{Proceedings of the ACM Web Conference 2022}, page 1070–1079, 2022.
\newblock URL \url{https://doi.org/10.1145/3485447.3512156}.

\bibitem[Xia et~al.(2024)Xia, Kao, and Huang]{llmgnn2}
Lianghao Xia, Ben Kao, and Chao Huang.
\newblock Opengraph: Towards open graph foundation models.
\newblock \emph{arXiv preprint arXiv:2403.01121}, 2024.

\bibitem[Xu et~al.(2019)Xu, Hu, Leskovec, and Jegelka]{gnn_work_4}
Keyulu Xu, Weihua Hu, Jure Leskovec, and Stefanie Jegelka.
\newblock How powerful are graph neural networks?
\newblock In \emph{International Conference on Learning Representations}, 2019.
\newblock URL \url{https://openreview.net/forum?id=ryGs6iA5Km}.

\bibitem[Yan et~al.(2023)Yan, Li, Long, Yan, Zhao, Zhuang, Yin, Zhang, Han, Sun, Deng, Zhang, Sun, Xie, and Wang]{TAG}
Hao Yan, Chaozhuo Li, Ruosong Long, Chao Yan, Jianan Zhao, Wenwen Zhuang, Jun Yin, Peiyan Zhang, Weihao Han, Hao Sun, Weiwei Deng, Qi~Zhang, Lichao Sun, Xing Xie, and Senzhang Wang.
\newblock A comprehensive study on text-attributed graphs: Benchmarking and rethinking.
\newblock In \emph{Thirty-seventh Conference on Neural Information Processing Systems Datasets and Benchmarks Track}, 2023.
\newblock URL \url{https://openreview.net/forum?id=m2mbfoSuJ1}.

\bibitem[Yang et~al.(2022)Yang, Wu, and Yan]{GKD}
Chenxiao Yang, Qitian Wu, and Junchi Yan.
\newblock Geometric knowledge distillation: Topology compression for graph neural networks.
\newblock In \emph{Advances in Neural Information Processing Systems}, 2022.
\newblock URL \url{https://openreview.net/forum?id=7WGNT3MHyBm}.

\bibitem[Ye et~al.(2023)Ye, Zhang, Wang, Xu, and Zhang]{natureisneed}
Ruosong Ye, Caiqi Zhang, Runhui Wang, Shuyuan Xu, and Yongfeng Zhang.
\newblock Natural language is all a graph needs.
\newblock \emph{arXiv preprint arXiv:2308.07134}, 2023.

\bibitem[Ying et~al.(2021{\natexlab{a}})Ying, Cai, Luo, Zheng, Ke, He, Shen, and Liu]{GraphCL}
Chengxuan Ying, Tianle Cai, Shengjie Luo, Shuxin Zheng, Guolin Ke, Di~He, Yanming Shen, and Tie-Yan Liu.
\newblock Do transformers really perform badly for graph representation?
\newblock In \emph{Advances in Neural Information Processing Systems}, pages 28877--28888, 2021{\natexlab{a}}.
\newblock URL \url{https://proceedings.neurips.cc/paper_files/paper/2021/file/f1c1592588411002af340cbaedd6fc33-Paper.pdf}.

\bibitem[Ying et~al.(2021{\natexlab{b}})Ying, Cai, Luo, Zheng, Ke, He, Shen, and Liu]{gtf2}
Chengxuan Ying, Tianle Cai, Shengjie Luo, Shuxin Zheng, Guolin Ke, Di~He, Yanming Shen, and Tie-Yan Liu.
\newblock Do transformers really perform badly for graph representation?
\newblock In \emph{Advances in Neural Information Processing Systems}, pages 28877--28888, 2021{\natexlab{b}}.
\newblock URL \url{https://proceedings.neurips.cc/paper_files/paper/2021/file/f1c1592588411002af340cbaedd6fc33-Paper.pdf}.

\bibitem[You et~al.(2020)You, Chen, Wang, and Shen]{gnn_work_7}
Y.~You, T.~Chen, Z.~Wang, and Y.~Shen.
\newblock L2-gcn: Layer-wise and learned efficient training of graph convolutional networks.
\newblock In \emph{2020 IEEE/CVF Conference on Computer Vision and Pattern Recognition (CVPR)}, pages 2124--2132, 2020.
\newblock URL \url{https://doi.ieeecomputersociety.org/10.1109/CVPR42600.2020.00220}.

\bibitem[You et~al.(2021)You, Chen, Shen, and Wang]{JOAO}
Yuning You, Tianlong Chen, Yang Shen, and Zhangyang Wang.
\newblock Graph contrastive learning automated.
\newblock In \emph{ICML}, 2021.
\newblock URL \url{https://arxiv.org/abs/2106.07594}.

\bibitem[Yu et~al.(2023)Yu, Ren, Gong, Tan, Li, and Zhang]{llmgnn1}
Jianxiang Yu, Yuxiang Ren, Chenghua Gong, Jiaqi Tan, Xiang Li, and Xuecang Zhang.
\newblock Empower text-attributed graphs learning with large language models (llms).
\newblock \emph{arXiv preprint arXiv:2310.09872}, 2023.

\bibitem[Yun et~al.(2019)Yun, Jeong, Kim, Kang, and Kim]{gtf1}
Seongjun Yun, Minbyul Jeong, Raehyun Kim, Jaewoo Kang, and Hyunwoo~J Kim.
\newblock Graph transformer networks.
\newblock In \emph{Advances in Neural Information Processing Systems}, 2019.
\newblock URL \url{https://proceedings.neurips.cc/paper_files/paper/2019/file/9d63484abb477c97640154d40595a3bb-Paper.pdf}.

\bibitem[Zhang et~al.(2024)Zhang, Sun, Wang, Fan, Mo, Xu, Liu, Yang, and Shi]{graphtranslator}
Mengmei Zhang, Mingwei Sun, Peng Wang, Shen Fan, Yanhu Mo, Xiaoxiao Xu, Hong Liu, Cheng Yang, and Chuan Shi.
\newblock Graphtranslator: Aligning graph model to large language model for open-ended tasks.
\newblock In \emph{Proceedings of the ACM Web Conference 2023}, 2024.

\bibitem[Zhang et~al.(2022)Zhang, Liu, Sun, and Shah]{GLNN}
Shichang Zhang, Yozen Liu, Yizhou Sun, and Neil Shah.
\newblock Graph-less neural networks: Teaching old {MLP}s new tricks via distillation.
\newblock In \emph{International Conference on Learning Representations}, 2022.
\newblock URL \url{https://openreview.net/forum?id=4p6_5HBWPCw}.

\bibitem[Zhu et~al.(2020)Zhu, Xu, Yu, Liu, Wu, and Wang]{zhu2020deep}
Yanqiao Zhu, Yichen Xu, Feng Yu, Qiang Liu, Shu Wu, and Liang Wang.
\newblock Deep graph contrastive representation learning.
\newblock \emph{arXiv preprint arXiv:2006.04131}, 2020.

\bibitem[Zhu et~al.(2021)Zhu, Xu, Yu, Liu, Wu, and Wang]{GCA}
Yanqiao Zhu, Yichen Xu, Feng Yu, Qiang Liu, Shu Wu, and Liang Wang.
\newblock Graph contrastive learning with adaptive augmentation.
\newblock In \emph{Proceedings of the Web Conference 2021}, page 2069–2080, 2021.
\newblock URL \url{https://doi.org/10.1145/3442381.3449802}.

\end{thebibliography}

\newpage
\appendix

\section{Dataset description}
\label{sec:data_stats}

\begin{table}[H]
    \centering
    \caption{Dataset statistics}
    \label{dataset_table}
    \resizebox{0.6\textwidth}{!}{%
    \begin{tabular}{llrrr}
    \toprule
    \textbf{Domain} & \textbf{Dataset} & \textbf{\#Nodes} & \textbf{\#Edges} & \textbf{\#Classes}  \\
    \midrule
    \multirow{3}{*}{Citation} & Arxiv & 169,343 & 1,166,243 & 40 \\
                              & Pubmed & 19,717 & 44,338 & 3 \\
                              & Cora & 25,120 & 91,140 & 70 \\
    \midrule
    \multirow{5}{*}{E-commerce} & Ele-Computer & 87,229 & 721,081 & 10 \\
                                & Ele-Photo & 48,362 & 500,928 & 12 \\
                                & Book-Children & 76,875 & 1,554,578 & 24 \\
                                & Book-History & 41,551 & 358,574 & 12 \\
                                & Sports-Fitness & 173,055 & 1,773,500 & 13 \\
    \bottomrule
    \end{tabular}}
\end{table}

\paragraph{Citation datasets}
The Arxiv dataset~\cite{OGB} represents a directed citation network among Computer Science (CS) papers from the arXiv preprint server. Each node in this graph corresponds to a paper, while edges represent citation links. The PubMed dataset~\cite{Pubmed} comprises 19,717 scientific publications from the PubMed database related to diabetes, which are categorized into three distinct classes: Experimentally induced diabetes, Type 1 diabetes, and Type 2 diabetes. This classification reflects the focus of each publication within the broader context of diabetes research. Lastly, the Cora dataset~\cite{Cora}, formally known as the ``Cora Research Paper Classification Dataset'', provides a comprehensive network for analyzing research paper classifications in machine learning. It is an extended version of the dataset commonly referred to in other studies~\cite{cora_old}, featuring detailed categorizations.

\paragraph{E-commmerce datasets}
All e-commerce datasets are provided in the TAG benchmark~\cite{TAG}. The Books-Children and Books-History datasets are extracted from the Amazon-Books dataset. Books-Children includes items with the second-level label ``Children'', while Books-History includes items with the second-level label ``History''. Each dataset's label corresponds to the three-level label of the book. The Ele-Computers dataset comprises items with the second-level label ``Computers'', and Ele-Photo includes items with the second-level label ``Photo''. Each of these datasets is labeled at the third level for electronic products. The Sports-Fitness dataset, sourced from the Amazon-Sports dataset, contains items with the second-level label ``Fitness''. Nodes in this dataset represent fitness-related items, and an edge between two items indicates they are frequently co-purchased or co-viewed.

\section{More experimental results}

\subsection{Legality rate}
\label{sec:legality}


\begin{table}[H]
    \centering
    \caption{Legality rate of LLM-backbone model (The worst results are marked in \colorbox{gray!30}{gray})}
    \resizebox{0.85\textwidth}{!}{%
    \begin{tabular}{c|cccccccc}
    \toprule
    \textbf{Dataset} & \textbf{Arxiv}  & \textbf{Computer} & \textbf{Pubmed} & \textbf{Cora} & \textbf{Children} & \textbf{History} & \textbf{Photo} & \textbf{Sports} \\
    \midrule
    Model & \multicolumn{8}{c}{Legality rate(\%)} \\
    \midrule
    Vicuna-7B-v1.5 & \cellcolor{gray!30}99.3 & \cellcolor{gray!30}96.7 & 100.0 & 95.8 & 99.2 & 98.9 & 94.1 & 99.6 \\
    LLaGA & 100.0 & 100.0 & \cellcolor{gray!30}98.9 & \cellcolor{gray!30}79.9 & \cellcolor{gray!30}93.1 & \cellcolor{gray!30}92.4 & \cellcolor{gray!30}77.8 & \cellcolor{gray!30}94.3\\
    \textbf{TEA-GLM} & 100.0 & 100.0 & 100.0 & 92.6 & 97.0 & 99.6 & 99.2 & 98.5 \\
    \bottomrule
    \end{tabular}
    \label{legality_rate}
    }
\end{table}

After training on specific datasets or tasks, large language models (LLMs) may produce invalid or incorrect answers to given questions. For instance, when handling unseen datasets or tasks, LLMs may generate responses that fall outside the set of acceptable answer candidates. To evaluate the impact of the training process on LLM performance, we follow the approach in \cite{graphtranslator} and use the legality rate to measure the proportion of valid answers produced by the model.

Table~\ref{legality_rate} demonstrates that the illegality rate of the LLaGA model significantly increases when exposed to datasets it has not previously encountered, suggesting a substantial impact of training methodologies on both the acquisition of knowledge and the model’s ability to generalize. Conversely, our model exhibits a notably stable performance across diverse unseen datasets, achieving higher legality rates in several cases.

\subsection{F1 score on node classification task}
\label{marco_f1}

\begin{table}[H]
    \centering
    \caption{Macro F1 of node classification task (\textbf{bold} highlights the best result across all methods, while \underline{underline} highlights the second-best results)}
    \label{citation_nc_f1}
    \Large
    \resizebox{\textwidth}{!}{%
    \begin{tabular}{cc|cc|cccc}
    \toprule
    \multirow{2}{*}{\textbf{Model type}} & \multirow{2}{*}{\textbf{Model}} & \multicolumn{2}{c|}{\textbf{Citation}} & \multicolumn{4}{c}{\textbf{E-commerce}} \\
    \cmidrule{3-8}
    & & \textbf{Pubmed} & \textbf{Cora} & \textbf{Children} & \textbf{History} & \textbf{Photo} & \textbf{Sports} \\
    \midrule
    ~ & MLP & 0.246$\pm$0.042 & 0.009$\pm$0.004 & 0.007$\pm$0.007 & 0.023$\pm$0.008 & 0.041$\pm$0.023 & 0.019$\pm$0.005 \\
    \midrule
    \multirow{9}{*}{\shortstack{GNN as\\ predictor}} & GCN & 0.187$\pm$0.021 & 0.007$\pm$0.001 & 0.006$\pm$0.004 & 0.024$\pm$0.013 & 0.034$\pm$0.007 & 0.017$\pm$0.009 \\
    & GraphSAGE & 0.257$\pm$0.084 & 0.007$\pm$0.003 & 0.005$\pm$0.003 & 0.029$\pm$0.024 & 0.020$\pm$0.011 & 0.021$\pm$0.004 \\
    & GAT & 0.259$\pm$0.065 & 0.006$\pm$0.001 & 0.063$\pm$0.067 & 0.159$\pm$0.117 & 0.036$\pm$0.035 & 0.091$\pm$0.090 \\
    & DGI & 0.213$\pm$0.127 & 0.004$\pm$0.002 & 0.012$\pm$0.004 & 0.038$\pm$0.015 & 0.045$\pm$0.015 & 0.018$\pm$0.005 \\
    & GKD & 0.247$\pm$0.039 & 0.004$\pm$0.001 & 0.028$\pm$0.003 & 0.060$\pm$0.008 & 0.049$\pm$0.015 & 0.050$\pm$0.008 \\
    & GLNN & 0.221$\pm$0.033 & 0.006$\pm$0.001 & 0.021$\pm$0.003 & 0.064$\pm$0.007 & 0.057$\pm$0.002 & 0.052$\pm$0.003 \\
    & NodeFormer & 0.232$\pm$0.089 & 0.008$\pm$0.003 & 0.019$\pm$0.008 & 0.046$\pm$0.031 & 0.055$\pm$0.006 & 0.049$\pm$0.009 \\
    & DIFFormer & 0.187$\pm$0.007 & 0.007$\pm$0.002 & 0.002$\pm$0.002 & 0.050$\pm$0.019 & 0.069$\pm$0.010 & 0.045$\pm$0.007 \\
    & OFA & 0.287$\pm$0.059 & 0.091$\pm$0.013 & 0.017$\pm$0.010 & 0.026$\pm$0.007 & 0.103$\pm$0.007 & 0.043$\pm$0.021 \\
    \midrule
    \multirow{5}{*}{\shortstack{LLM as\\ predictor}} & Vicuna-7B-v1.5 & 0.629$\pm$0.024 & 0.109$\pm$0.002 & \textbf{0.279$\pm$0.002} & \underline{0.349$\pm$0.003} & \underline{0.383$\pm$0.001} & 0.410$\pm$0.002 \\
    & GraphGPT-std & 0.649 & 0.082 & - & - & - & - \\
    & GraphGPT-cot & 0.482 & \underline{0.127} & - & - & - & - \\
    & LLaGA & \underline{0.778$\pm$0.056} & 0.108$\pm$0.014 & 0.163$\pm$0.029 & 0.144$\pm$0.025 & 0.362$\pm$0.039 & \textbf{0.446$\pm$0.035} \\
    & \textbf{TEA-GLM} & \textbf{0.839$\pm$0.012} & \textbf{0.148$\pm$0.015} & \underline{0.252$\pm$0.005} & \textbf{0.365$\pm$0.011} & \textbf{0.421$\pm$0.032} & \underline{0.430$\pm$0.009} \\
    \bottomrule
    \end{tabular}}
\end{table}

Due to the absence of a metric to calculate the F1 score while considering the illegality rate, we adopt the methodology used in~\cite{graphtranslator}. For the LLM-backbone models, we only calculate the Macro F1 score for legally permissible responses provided by the model. This calculation method may not accurately reflect the model's performance fully. Therefore, we also report the illegality rate in Table~\ref{legality_rate}. Please note that the accuracy metric is unaffected by illegal responses, which are considered error responses.

\subsection{Supervised results}
\label{supervised}

\begin{table}[H]
    \centering
    \caption{Accuracy and macro F1 on training datasets (\textbf{bold} highlights the best result across all methods, while \underline{underline} highlights the second-best results)}
    \label{supervised_learning_table}
    \Large
    \resizebox{0.8\textwidth}{!}{%
    \begin{tabular}{cc|cc|cc}
    \toprule
    \multirow{2}{*}{\textbf{Model type}} & \multirow{2}{*}{\textbf{Model}} & \multicolumn{2}{c|}{\textbf{Arxiv}} & \multicolumn{2}{c}{\textbf{Computer}} \\
    \cmidrule{3-6}
    \multirow{2}{*}{} & \multirow{2}{*}{} & Acc & F1 & Acc & F1 \\
    \midrule
    & MLP & 0.546$\pm$0.004 & 0.295$\pm$0.007 & 0.420$\pm$0.006 & 0.267$\pm$0.005  \\
    \midrule
    \multirow{9}{*}{\shortstack{GNN as\\ predictor}} & GCN & 0.545$\pm$0.005 & 0.317$\pm$0.006 & 0.424$\pm$0.012 & 0.386$\pm$0.014  \\
    & GraphSAGE & 0.556$\pm$0.006 & 0.315$\pm$0.008 & 0.534$\pm$0.037 & 0.347$\pm$0.036  \\
    & GAT & 0.561$\pm$0.003 & 0.339$\pm$0.005 & 0.609$\pm$0.035 & \underline{0.598$\pm$0.039}  \\
    & DGI & 0.342$\pm$0.024 & 0.336$\pm$0.011 & 0.594$\pm$0.004 & 0.452$\pm$0.008  \\
    & GKD & 0.393$\pm$0.085 & 0.164$\pm$0.029 & 0.351$\pm$0.031 & 0.155$\pm$0.016  \\
    & GLNN & 0.602$\pm$0.004 & 0.362$\pm$0.008 & 0.393$\pm$0.005 & 0.243$\pm$0.007  \\
    & NodeFormer & 0.544$\pm$0.016 & 0.297$\pm$0.029 & 0.434$\pm$0.012 & 0.288$\pm$0.012  \\
    & DIFFormer & 0.616$\pm$0.025 & 0.356$\pm$0.024 & 0.629$\pm$0.012 & 0.467$\pm$0.022  \\
    & OFA & \underline{0.682$\pm$0.006} & \underline{0.495$\pm$0.006} & \textbf{0.753$\pm$0.004} & \textbf{0.687$\pm$0.006}  \\
    \midrule
    \multirow{5}{*}{\shortstack{LLM as\\ predictor}} & Vicuna-7B-v1.5 & 0.347$\pm$0.000 & 0.164$\pm$0.001 & 0.372$\pm$0.010 & 0.304$\pm$0.002 \\
    & GraphGPT-std & 0.626 & 0.262 & - & - \\
    & GraphGPT-cot & 0.576 & 0.228 & - & - \\
    & LLaGA & \textbf{0.749$\pm$0.001} & \textbf{0.575$\pm$0.003} & \underline{0.642$\pm$0.004} & 0.562$\pm$0.001 \\
    & \textbf{TEA-GLM} & 0.655$\pm$0.001 & 0.445$\pm$0.002 & 0.578$\pm$0.002 & 0.496$\pm$0.010 \\
    \bottomrule
    \end{tabular}
    }
\end{table}

We report the supervised learning results in Table~\ref{supervised_learning_table}. The GNN-backbone models continue to demonstrate robust performance in fitting training data. Similarly, the LLaGA model shows its efficacy in supervised learning scenarios. However, despite their strong performance on training datasets, these models exhibit limited generalization capabilities on unseen datasets as shown in Table~\ref{tab:nc_citation} and Table~\ref{citation_nc_f1}.

\section{Parameter sensitivity analysis}
\label{Sensitivity}
\begin{figure}[H]
    \centering
    \subfigure[Arxiv]{
    \label{token_arxiv}
        \includegraphics[scale=0.17]{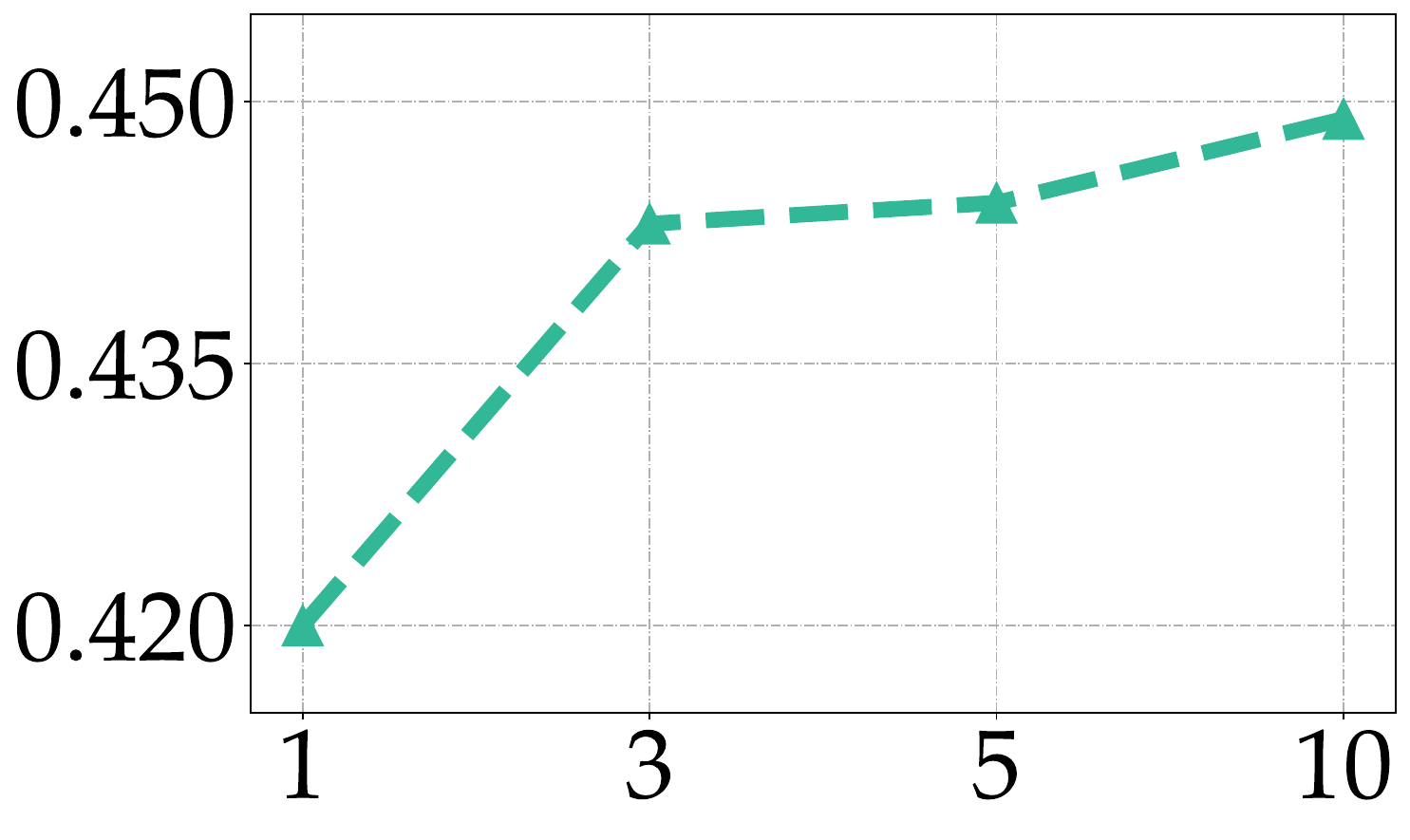}
    }
    \subfigure[Pubmed]{
    \label{token_pubmed}
        \includegraphics[scale=0.17]{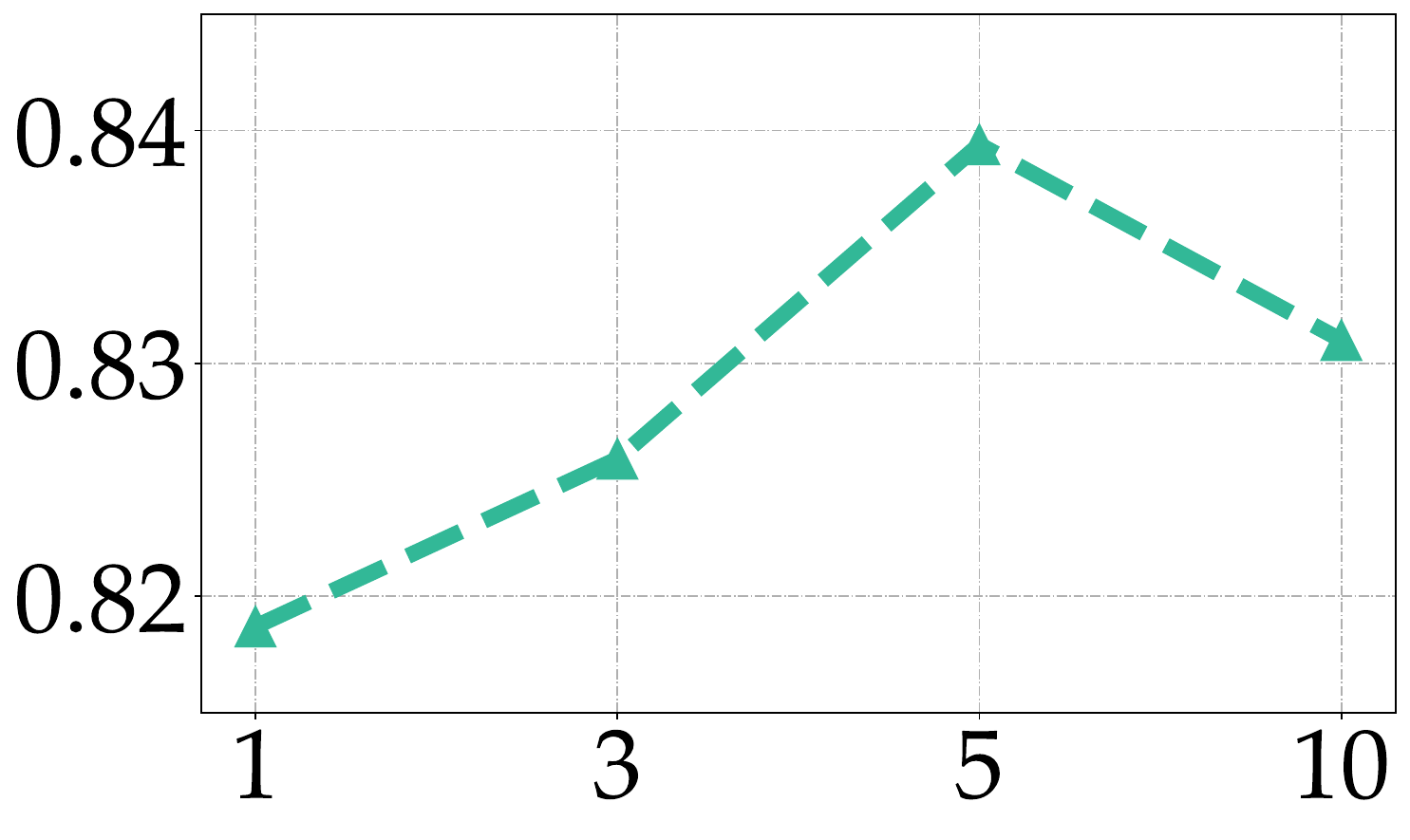}
    }
    \subfigure[Cora]{
    \label{token_cora}
        \includegraphics[scale=0.17]{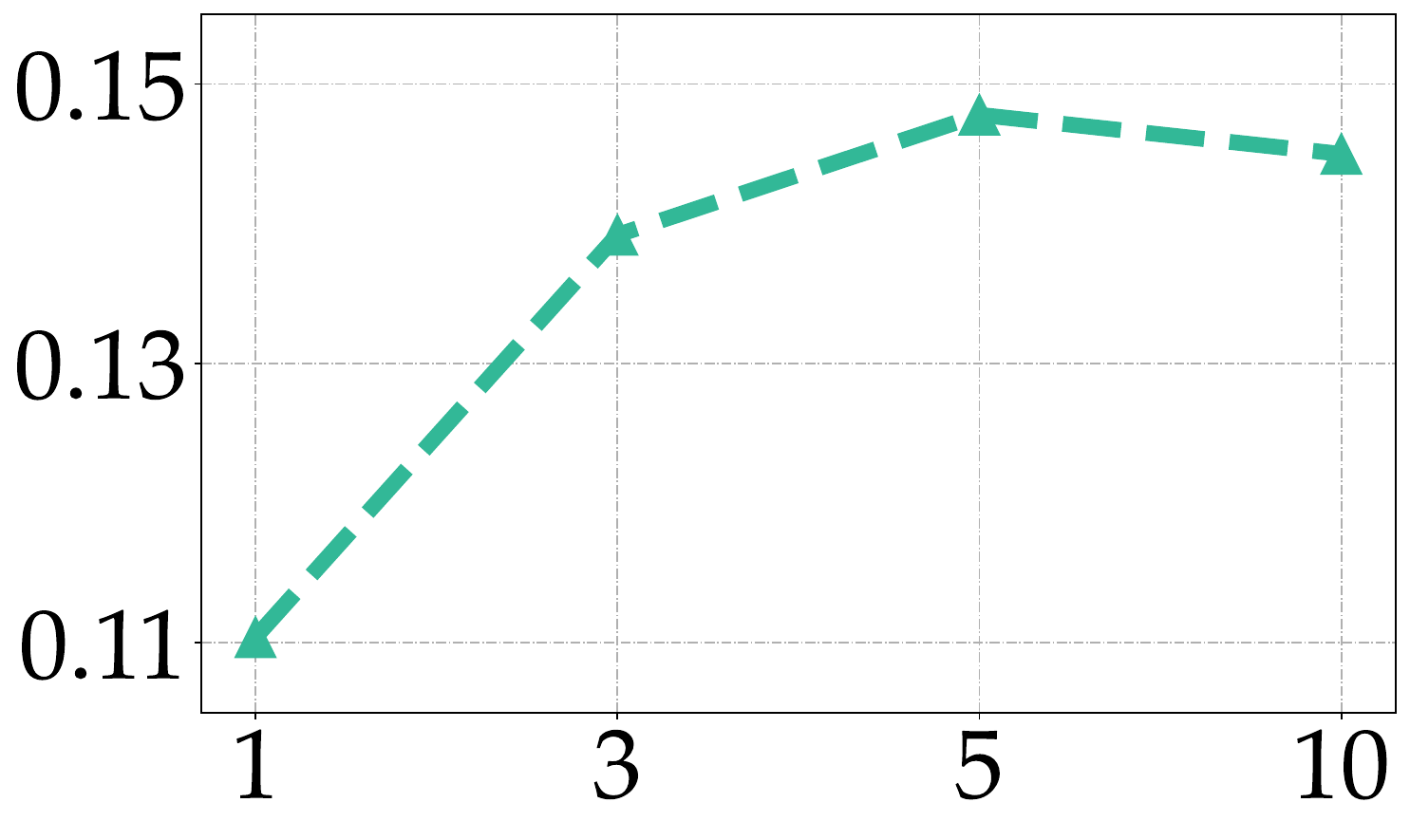}
    }
    \caption{Impact of number of graph token embeddings~(Macro F1)}
    \label{sensitive_token}
\end{figure}

\begin{figure}[H]
    \centering
    \subfigure[Arxiv]{
    \label{tp_arxiv}
        \includegraphics[scale=0.17]{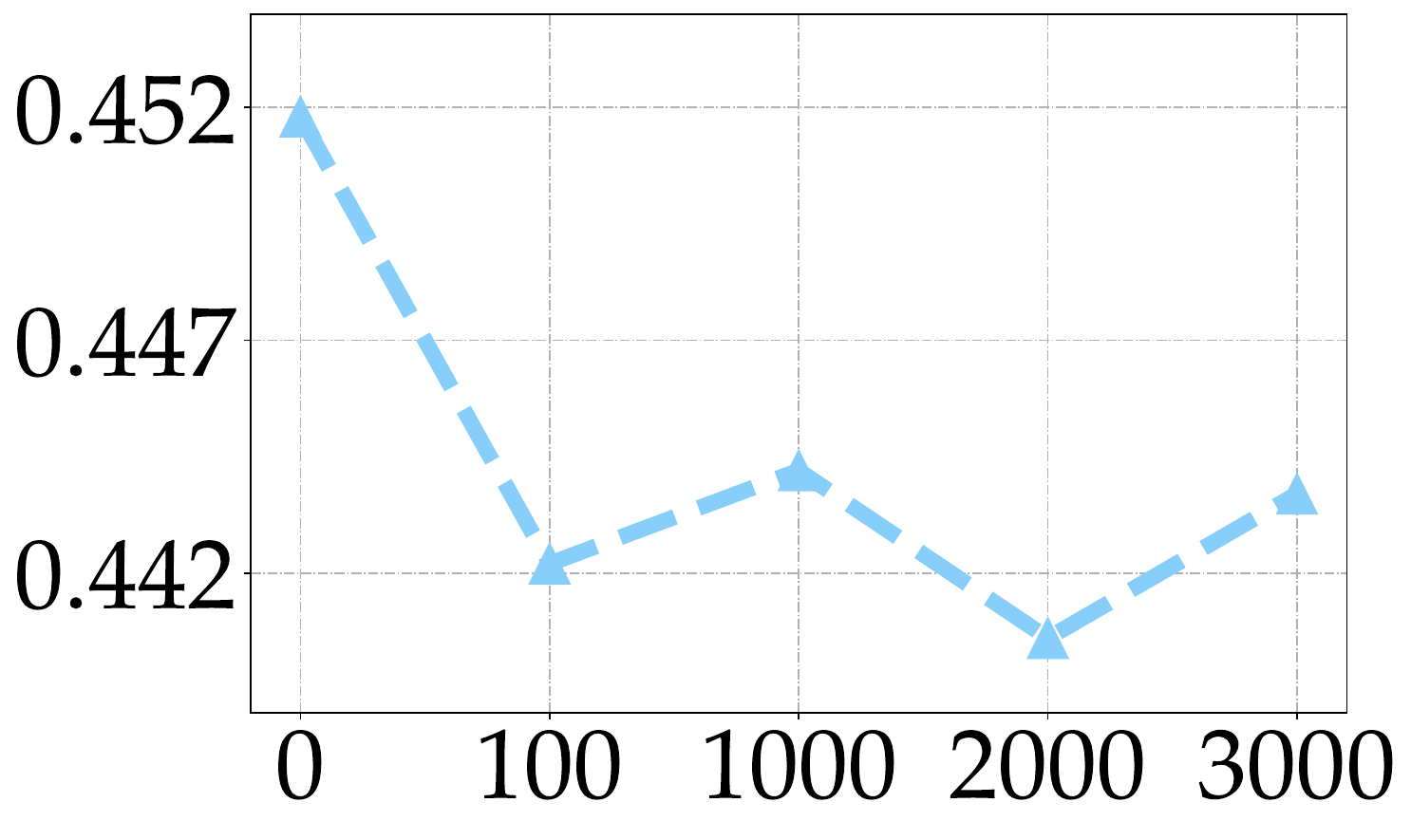}
    }
    \subfigure[Pubmed]{
    \label{tp_pubmed}
        \includegraphics[scale=0.17]{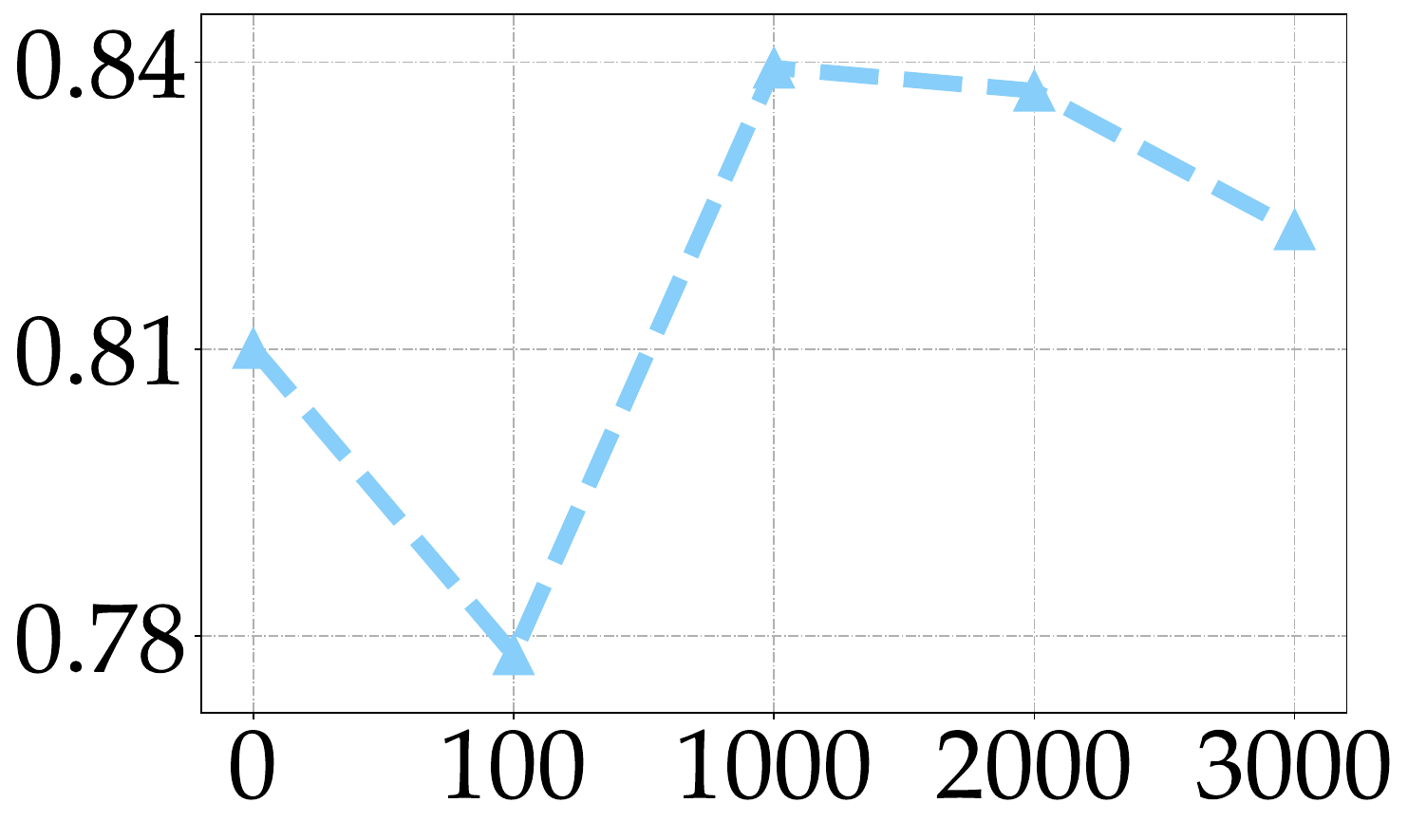}
    }
    \subfigure[Cora]{
    \label{tp_cora}
        \includegraphics[scale=0.17]{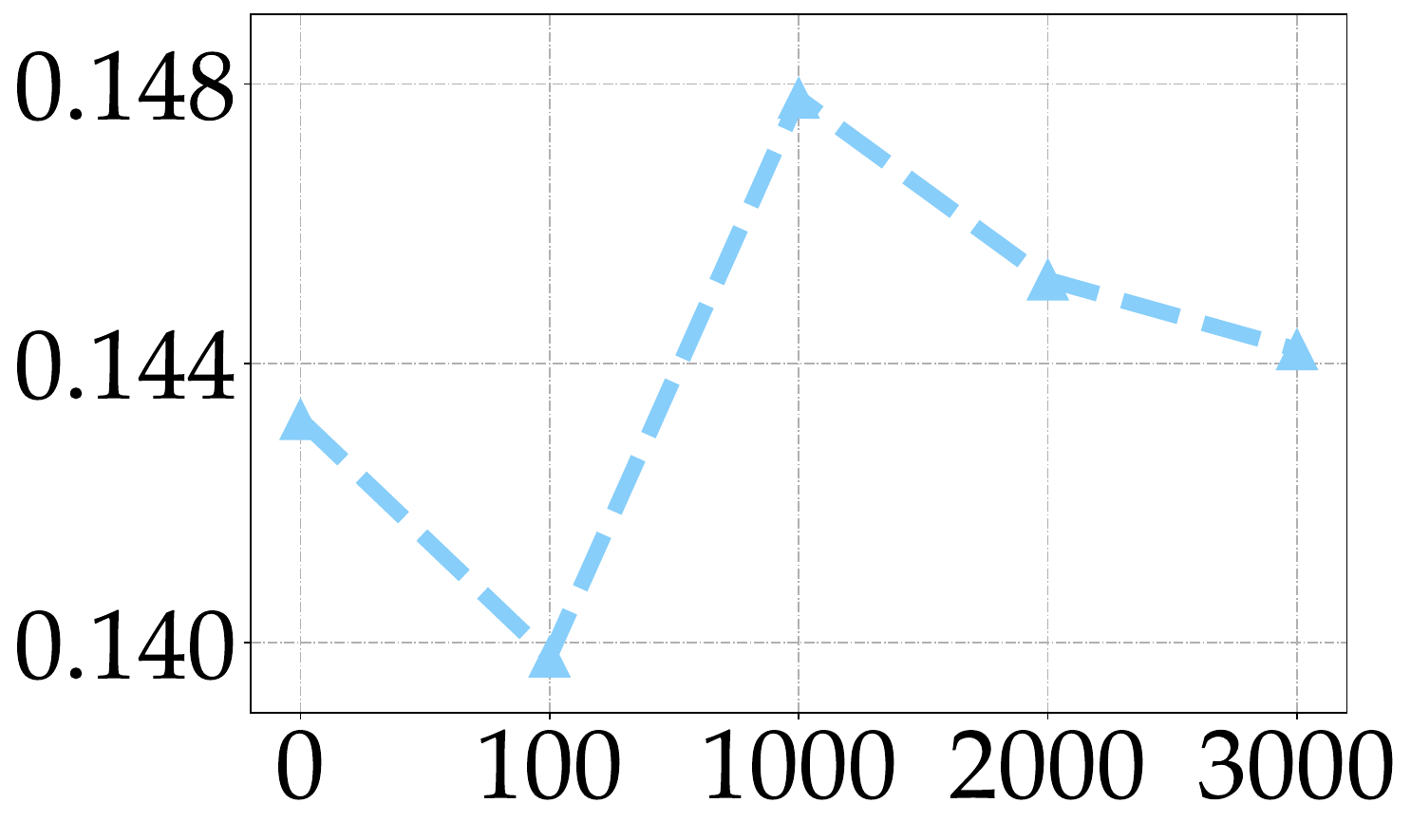}
    }
    \caption{Impact of number of principal components~(Macro F1)}
    \label{sensitive_tp}
\end{figure}

\paragraph{Number of graph token embeddings}

To discuss the impact of the number of graph token embeddings, we set $K \in \left\{ 1, 3, 5, 10\right\}$ and report the results on node classification task in Figure~\ref{sensitive_token}. In the context of training datasets and unseen datasets, we observe two distinct patterns. With an increase in the number of graph token embeddings in the training dataset, there is a slight improvement in the model's performance on that dataset. This suggests that in a supervised learning scenario, enhancing the model's performance can be achieved by increasing the quantity of graph token embeddings. Conversely, for unseen datasets, our model requires only a minimal number of graph token embeddings to achieve satisfactory performance, indicating that the number of learnable parameters in our model is significantly less than concurrent works.

\paragraph{Number of principal components}

We define $P \in \left\{0, 100, 1000, 2000, 3000\right\}$ and discuss the results of the node classification task in Figure~\ref{sensitive_tp}. In supervised learning scenarios, omitting contrastive learning with principal components can lead to a slight increase in accuracy. However, this often makes the model more prone to overfitting on training datasets. When the number of principal components is too small, it adversely affects the model's learning capability. Remarkably, when $P=1000$, the model demonstrates satisfactory performance. At this level, the principal components capture $50\%$ of the variance of LLM's token embeddings.
 
\section{Complete instructions}
\label{sec:instruction}

\begin{figure}[H]
    \centering
    \includegraphics[scale=0.45]{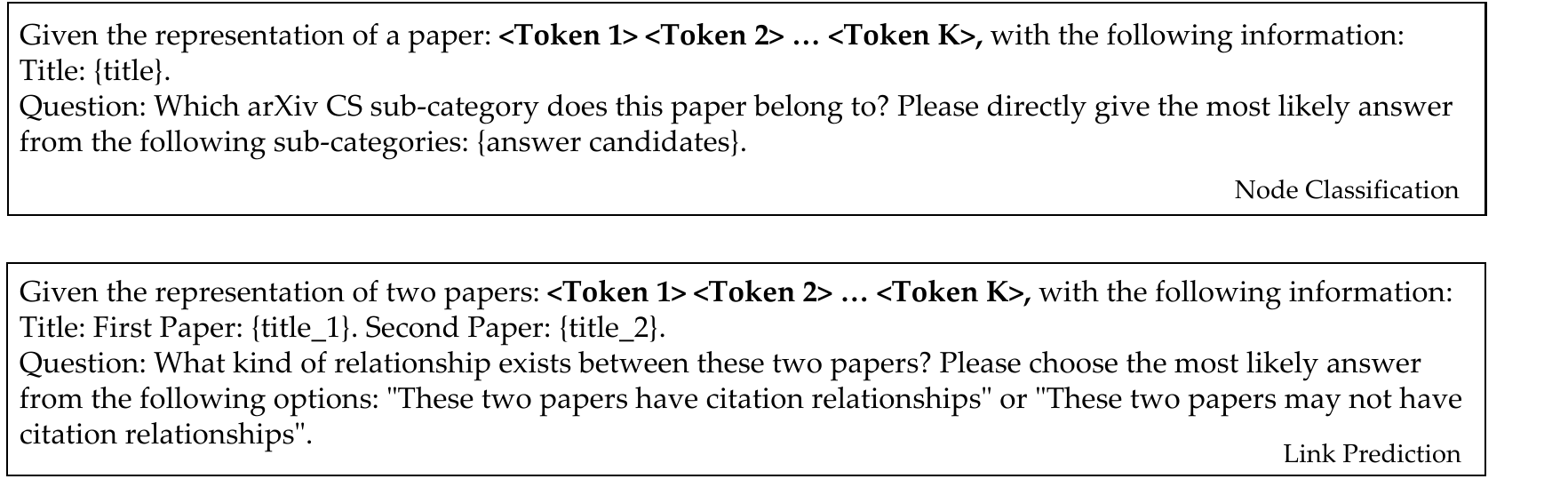}
    \caption{Instructions for node classification and link prediction}
    \label{fig:instruction}
\end{figure}

In node classification tasks, we provide candidate labels to facilitate the model's learning process, focusing on discovering the correct answers rather than merely memorizing them. For link prediction, we structure the instructions in a format similar to that of node classification. This approach is designed to enhance the model's ability to transfer learned knowledge effectively across different tasks.

\section{Cross-task zero-shot results with different pooling methods}

\begin{table}[H]
    \centering
    \caption{AUC of link prediction~(Cross-task) with different pooling methods}
    \label{cross_task_pooling}
    \resizebox{0.5\textwidth}{!}{%
    \begin{tabular}{c|ccc}
    \toprule
    \multirow{2}{*}{\textbf{Model}} & \multicolumn{3}{c}{\textbf{Citation}}  \\
    \cmidrule{2-4}
    & \textbf{Arxiv} & \textbf{Pubmed} & \textbf{Cora} \\
    \midrule
    OFA & 0.469 & 0.481 & 0.492 \\
    Vicuna-7B-v1.5 & 0.513 & 0.543 & 0.527 \\
    Vicuna-7B-SPT & 0.537 & 0.535 & 0.565  \\
    GraphGPT-std & 0.649 & 0.501 & 0.520 \\
    LLaGA & 0.570 & 0.569 & 0.537 \\
    \midrule
    TEA-GLM (max) & 0.639 & 0.650 & 0.566 \\
    TEA-GLM (sum) & 0.657 & 0.689 & 0.586 \\
    \textbf{TEA-GLM (mean)} & \textbf{0.659} & \textbf{0.690} & \textbf{0.588} \\
    \bottomrule
    \end{tabular}
    }
\end{table}

Considering that different pooling methods may impact cross-task performance, we conducted experiments using three common pooling methods separately, and the results are shown in the Table \ref{cross_task_pooling}.

\end{document}